\documentclass[10pt,twocolumn,letterpaper]{article}

\usepackage[pagenumbers]{iccv} %
\usepackage{graphicx}
\usepackage{booktabs}
\usepackage{multirow}
\usepackage{amssymb}
\usepackage{tabularx}
\usepackage{tabularray}
\usepackage{subcaption}
\usepackage[hang, flushmargin]{footmisc}
\usepackage{array}
\newcolumntype{C}{>{\centering \arraybackslash}X}
\usepackage[accsupp]{axessibility}  %

\newcommand{\gmean}{\boldsymbol{\mu}}
\newcommand{\gcovariance}{\mathbf{\Sigma}}
\newcommand{\grot}{\mathbf{R}}
\newcommand{\gscale}{\mathbf{s}}
\newcommand{\gopacity}{o}
\newcommand{\gcolor}{\mathbf{c}}

\newcommand{\motioncoef}{\mathbf{w}}
\newcommand{\SETHREE}{\mb{SE}(3)}
\newcommand{\ftf}[3]{{#1}_{#3 \rightarrow #2}}

\newcommand{\deltaavg}{<\hspace{-0.2em}\delta_{\text{avg}}}
\newcommand{\smallthreedth}{.05}
\newcommand{\largethreedth}{.10}
\newcommand{\deltathreedsmall}{\delta_{3D}^{\smallthreedth}}
\newcommand{\deltathreedlarge}{\delta_{3D}^{\largethreedth}}
\newcommand{\tf}[1]{\mathbf{#1}}

\newcommand{\world}{{}^{\textrm{w}}}
\newcommand{\cam}{{}^{\textrm{c}}}

\newcommand{\mb}[1]{\mathbb{#1}}
\newcommand{\mc}[1]{\mathcal{#1}}

\definecolor{iccvblue}{rgb}{0.21,0.49,0.74}
\usepackage[pagebackref,breaklinks,colorlinks,allcolors=iccvblue]{hyperref}

\def\paperID{9473} %
\def\confName{ICCV}
\def\confYear{2025}

\title{Shape of Motion: 4D Reconstruction from a Single Video}

\author{Qianqian Wang$^{1,2*}$, Vickie Ye$^{1*}$, Hang Gao$^{1*}$, Weijia Zeng$^{3*}$,
\\
Jake Austin$^{1}$, Zhengqi Li$^{4}$, Angjoo Kanazawa$^{1}$
\\
\vspace{-0.5em}
\\
{
$^{1}$UC Berkeley \quad
$^{2}$Google DeepMind \quad
$^{3}$UC San Diego \quad
$^{4}$Adobe Research
\vspace{-0.1em}
}
}

\begin{document}
\twocolumn[{%
\renewcommand\twocolumn[1][]{#1}%
\maketitle
\vspace{-1em}
\begin{center}
\centering
\captionsetup{type=figure}
\includegraphics[width=\linewidth]{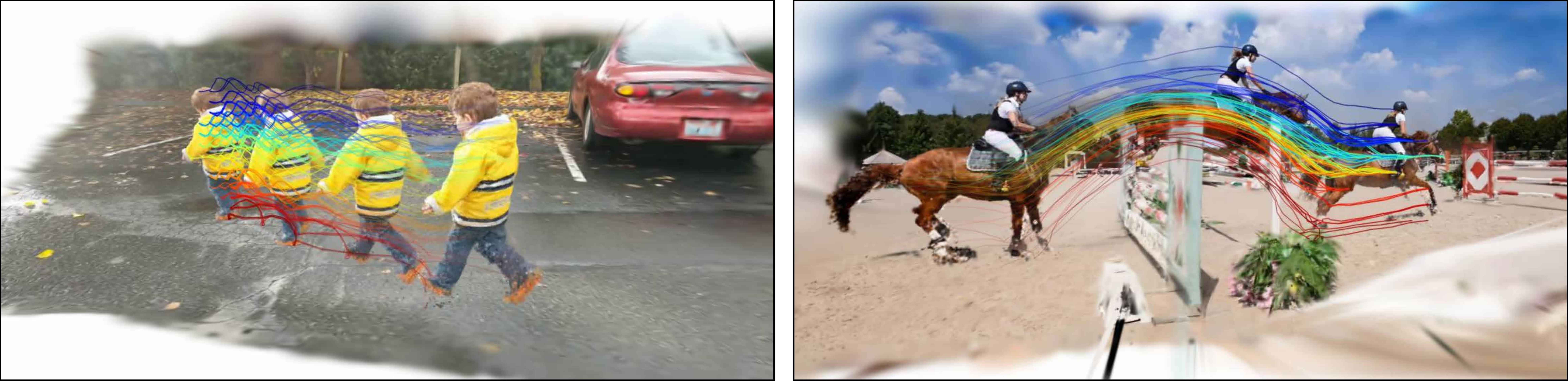}
\captionof{figure}{\textbf{Shape of Motion.} \small{
Our method enables joint long-range 3D tracking and novel view synthesis from a monocular video of a complex dynamic scene. We render moving elements at a fixed viewpoint across time and visualize estimated 3D motion as colorful trajectories. These trajectories reveal distinct geometric patterns, which leads to the term ``Shape of Motion''.
}
} 
\label{fig:teaser}
\end{center}

}]
\begin{abstract}
Monocular dynamic reconstruction is a challenging and long-standing vision problem due to the highly ill-posed nature of the task. Existing approaches depend on templates, are effective only in quasi-static scenes, or fail to model 3D motion explicitly. We introduce a method for reconstructing generic dynamic scenes, featuring explicit, persistent 3D motion trajectories in the world coordinate frame, from casually captured monocular videos.
We tackle the problem with two key insights: First, we exploit the low-dimensional structure of 3D motion by representing scene motion with a compact set of $\SETHREE$ motion bases. Each point's motion is expressed as a linear combination of these bases, facilitating soft decomposition of the scene into multiple rigidly-moving groups. Second, we 
take advantage of off-the-shelf data-driven priors such as monocular depth maps and long-range 2D tracks, and devise a method to effectively consolidate these noisy supervisory signals, resulting in a globally consistent representation of the dynamic scene. Experiments show that our method achieves state-of-the-art performance for both long-range 3D/2D motion estimation and novel view synthesis on dynamic scenes. Project Page: \url{https://shape-of-motion.github.io/}
\renewcommand{\thefootnote}{}
\footnotetext{$^*$Equal contribution}

\end{abstract}    
\section{Introduction}
\label{sec:intro}
Reconstructing the persistent geometry and their 3D motion across a video is crucial for understanding and interacting with the underlying physical world. While recent years have seen impressive progress in modeling static 3D scenes~\cite{mildenhall2020nerf, kerbl20233d}, recovering the geometry and motion of complex dynamic 3D scenes, especially from a single video, remains an open challenge.
A number of prior dynamic reconstruction and novel view synthesis approaches have attempted to tackle this problem.
However, most methods rely on synchronized multi-view videos~\cite{broxton2020immersive, fridovich2023k, fridovich2023k, Luiten2023Dynamic3G, wu20234d} or additional LIDAR/depth sensors~\cite{Gu2019HPLFlowNetHP, Wang2022NeuralPF, Liu2018LearningSF, Puy2020FLOTSF, Wang2019FlowNet3DGL}. 
Recent monocular approaches can operate on regular dynamic videos, but they typically model 3D scene motion as short-range scene flow between consecutive times~\cite{li2021neural, li2023dynibar, Gao-ICCV-DynNeRF} or deformation fields that map between canonical and view space~\cite{park2021nerfies, park2021hypernerf, yang2023deformable3dgs}, failing to capture 3D motion trajectories persistent over a video.

The longstanding challenge for general in-the-wild videos lies in the poorly constrained nature of the 4D reconstruction problem.
In this work, we tackle this challenge with two key insights.
The first is that, while the image space dynamics can be complex and discontinuous,
the underlying 3D motion is a composition of continuous simple
rigid motions.
Our second insight is that data-driven priors provide complementary, though noisy cues, that aggregate well into a globally coherent representation of the 3D scene geometry and motion.

Motivated by these two insights, we represent the dynamic scene as a set of persistent 3D Gaussians,
and represent their motion across the video in terms of a compact set of shared $\mb{SE}(3)$ motion bases. %
Unlike traditional scene flow, which computes 3D correspondences between consecutive frames, our representation recovers a persistent 3D trajectory over the whole video, enabling long-range 3D tracking across the entire video.
As the 3D trajectories produced by our method capture the geometric patterns that trace each point’s movement through 3D space and time as shown in Figure~\ref{fig:teaser}, we refer to our approach as ``Shape of Motion''.
We show how to fit our explicit scene representation to a general video in-the-wild, 
by fusing together complementary cues from two main sources: monocular depth estimates per-frame, and 2D track estimates across frames.
We conduct extensive evaluations on both synthetic and real-world dynamic video datasets, 
and show that our proposed approach significantly outperforms prior methods in both long-range 2D/3D tracking and novel view synthesis. 

In summary, our key contributions are:
(1) A 4D scene representation enabling both real-time novel view synthesis and globally consistent 3D tracking for any point at any time.
(2) An optimization framework that fuses learned motion and geometry priors into a unified 4D representation from a single, casually captured monocular video. %

\section{Related Work}
\label{sec:related}

\noindent\textbf{Correspondences and Tracking.}
While monocular 3D long range tracking remains largely unexplored, numerous approaches track in 2D image space, typically using optical flow for point correspondences. This involves estimating dense motion fields between image pairs~\cite{Horn1981DeterminingOF, black1993framework, Lucas1981AnII, Brox2009LargeDO, Brox2004HighAO, Sun2017PWCNetCF, Dosovitskiy2015FlowNetLO, huang2022flowformer, jiang2021learning, teed2020raft, ilg2017flownet, huang2022flowformer, xu2022gmflow, jiang2021learningf, janai2018unsupervised, ren2019fusion, shi2023videoflow}. 
Sparse keypoint matching methods can enable long trajectory generation~\cite{Bay2006SURFSU, Rublee2011ORBAE, Lowe2004DistinctiveIF, DeTone2017SuperPointSI, Liu2011SIFTFD}, but these methods are primarily intended for sparse 3D reconstruction.
Long-range 2D trajectory estimation for arbitrary points has been explored in earlier works, which relied on hand-crafted priors to generate motion trajectories~\cite{Rubinstein2012TowardsLL, birchfield2008joint, sandlong, sand2008particle, Sivic2004ObjectLG, Wang2013ActionRW}. Recently, there has been a resurgence of interest in this problem, with several works showcasing impressive long-range 2D tracking results on challenging, in-the-wild videos. These approaches employ either test-time optimization where models consolidate noisy short-range motion estimates into long-term correspondences~\cite{wang2023omnimotion, Neoral_2024_WACV,song2024track}, or data-driven strategies~\cite{harley2022particle, karaev2023cotracker, doersch2023tapir, doersch2024bootstap}, where neural networks learn long-term correspondence estimates from synthetic data~\cite{doersch2022tap, zheng2023pointodyssey}.
While these methods effectively track any 2D point throughout a video, they lack the knowledge of underlying 3D scene geometry and motions.

To model 3D scene motion, recent work in monocular settings has explored self-supervised learning and test-time optimization~\cite{NEURIPS2021_a11f9e53, Hur2020SelfSupervisedMS, li2021neural, li2023dynibar, Yang2021LASRLA, Yang2021BANMoBA}. More recent approaches estimate 3D trajectories in general scenes in a feedforward manner~\cite{xiao2024spatialtrackertracking2dpixels, ngo2025deltadenseefficientlongrange,koppula2024tapvid}, but the predicted motion remains in the \emph{frame} space, entangling object and camera motion.
In contrast, our method recovers persistent, long-range 3D trajectories in the \emph{world} coordinate space.

\vspace{0.2em}
\noindent\textbf{Dynamic Reconstruction and View Synthesis.}
Our work also relates to dynamic 3D scene reconstruction and novel view synthesis. In non-rigid reconstruction, early methods often required RGB-D sensors~\cite{Zollhfer2014RealtimeNR, Newcombe2015DynamicFusionRA, Bozic2019DeepDeformLN, Dou2016Fusion4D, Innmann2016VolumeDeformRV} or strong hand-crafted priors~\cite{russell2014video, Kumar2017MonocularD3, Ranftl_2016_CVPR}. A family of works~\cite{novotny2019c3dpocanonical3dpose, 10.1007/s11263-013-0684-2, 854941} apply low-rank assumptions to the underlying motion model. Recent work has demonstrated progress toward general dynamic reconstruction either by integrating monocular depth priors~\cite{li2019learning, zhang2021consistent, luo2020consistent, kopf2021robust, zhang2022structure,li2024_MegaSaM}, or directly learning to perform dynamic reconstruction~\cite{dust3r_cvpr24,zhang2024monst3r,lu2025align3r,wang2025continuous} from data. However, these methods do not perform long-range 3D tracking, and do not focus on novel view synthesis.

Neural Radiance Fields~(NeRF)~\cite{mildenhall2020nerf} and Gaussian Splatting~\cite{kerbl3Dgaussians}  have shown strong performance in novel view synthesis.
For dynamic scenes, many methods~\cite{bansal20204d, fridovich2023k, broxton2020immersive, li2022neural, fridovich2023k, cao2023hexplane, song2023nerfplayer, stich2008view, wang2022fourier,li2023spacetime} still require simultaneous multi-view video observations or predefined templates~\cite{isik2023humanrf, li2022tava, weng2022humannerf}. Template-free monocular approaches model dynamic scenes with different types of representations such as video depth maps~\cite{yoon2020novel}, time-dependent NeRFs~\cite{du2021neural, li2021neural, pumarola2021d, park2021nerfies, wang2021neural, park2021hypernerf, xian2021space, li2023dynibar}, and dynamic 3D Gaussians~\cite{wu20234d, duan20244d, yang2023real, yang2023deformable3dgs}.
Although significant progress has been made, as DyCheck~\cite{gao2022dynamic} pointed out, many approaches focus on scenarios with camera teleportation~\cite{kratimenos2024dynmf,yang2023deformable3dgs} or quasi-static scenes~\cite{park2021nerfies,park2021hypernerf}, which are effectively multi-view and do not represent real-world monocular videos. In this work, we focus on casually-captured monocular videos, a more practical and challenging setup. Recent and concurrent efforts~\cite{lei2024mosca,wang2024gflow,stearns2024dynamic,liu2024modgs,dreamscene4d} have also investigated similar settings, using strong off-the-shelf data-driven priors.

\begin{figure*}[t]
    \centering
    \includegraphics[width=.9\linewidth,trim={0 0cm
    0cm 0cm}]{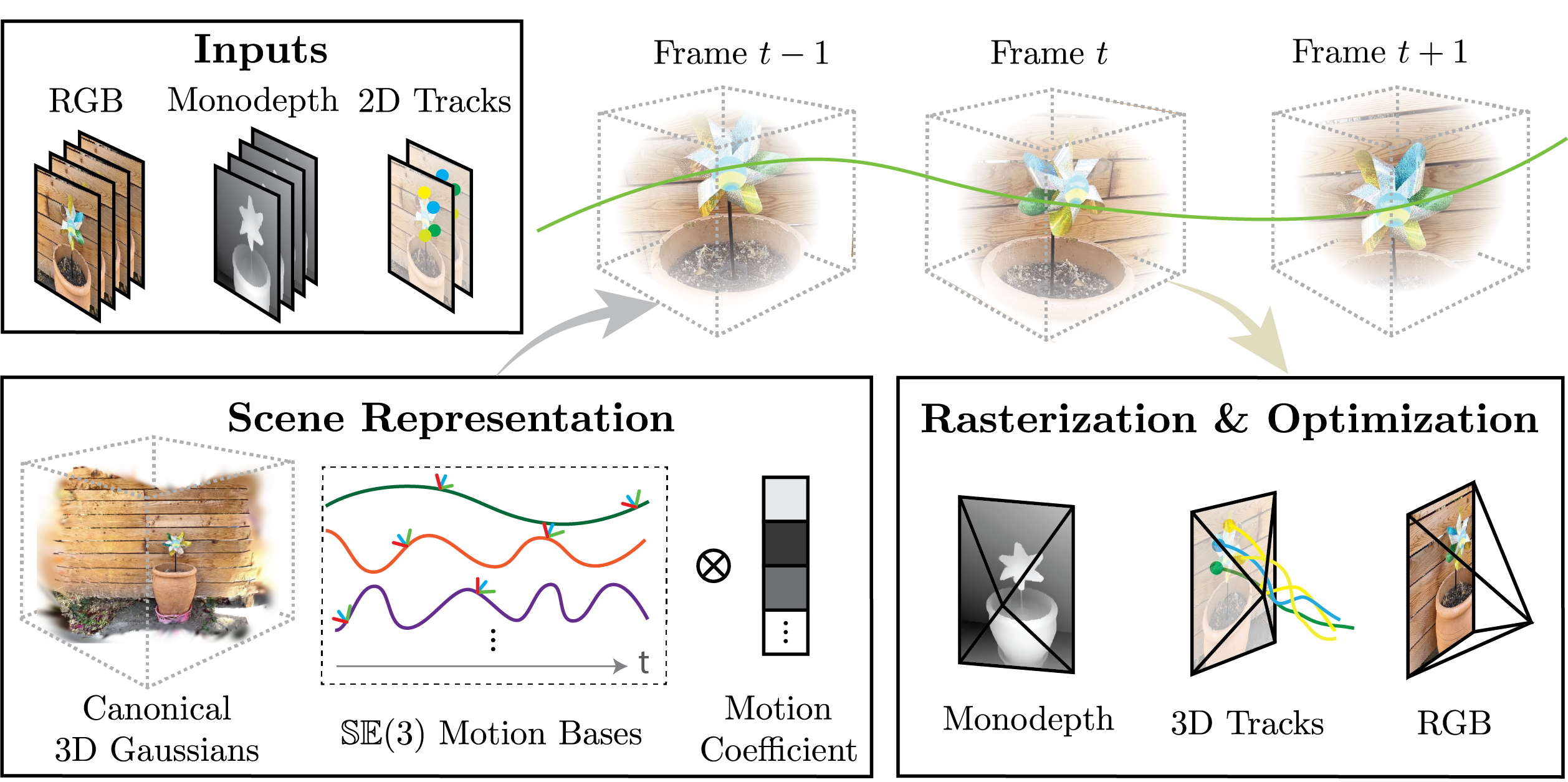}
    \caption{\textbf{System Overview.}
    Given a single RGB video sequence with known camera poses, monocular depth maps and 2D tracks computed from off-the-shelf models~\cite{depthanything, doersch2023tapir} as input,
    we optimize a dynamic scene representation as a set of persistent 3D Gaussians that translate and rotate over time. To capture the low-dimensional nature of scene motion, we model the motion with a set of compact $\SETHREE$ motion bases shared across all scene elements. Each 3D Gaussian's motion is represented as a linear combination of these %
    motion bases, weighted by motion coefficients specific to each Gaussian.
    We supervise our scene representation (canonical 3D Gaussians, per-Gaussian motion coefficients, and global motion bases) by comparing the rendered outputs (RGB, depths and 2D tracks) with the corresponding input signals. This results in a globally coherent dynamic 3D scene representation in the world coordinate with long-range 3D trajectories.    
    }
    \vspace{-1.5em}
    \label{fig:pipeline}
\end{figure*}

\section{Method}
\label{sec:method}
Our method takes as input a sequence of $T$ video frames $\{ I_t \in \mb{R}^{H \times W\times 3}  \}$ of a dynamic scene, the camera intrinsics $\tf K_t \in \mb{R}^{3\times 3}$, and world-to-camera extrinsics $\tf E_t \in \mb{SE}(3)$ of each input frame $I_t$.
From these inputs, we aim to recover the geometry of the entire dynamic scene and the full-length 3D motion trajectory of every point in the scene. 
Unlike most prior dynamic NeRFs methods~\cite{li2019learning, li2023dynibar, gao2022dynamic, xian2021space, wang2021neural} which render the scene contents through volumetric ray casting and represent the motion implicitly at fixed 3D locations,
we model the dense scene elements as a set of canonical 3D Gaussians, that  translate and rotate over entire video as persistent motion trajectories.
We adopt explicit point-based representation
because it simultaneously allows for both (1) high-fidelity rendering in real-time and (2) full-length 3D tracking of any surface point from any input time.

Optimizing an explicit representation of dynamic 3D Gaussians from a single video is severely ill-posed --- at each point in time, the moving subjects in the scene are observed  from only a single viewpoint.
In order to overcome this ambiguity, we make two insights:
First, 
while the projected 2D dynamics might be complex in the video, the underlying 3D motion in the scene is low-dimensional, and composed of simpler units of rigid motion.
Second, powerful data-driven priors, namely monocular depth estimates and long-range 2D tracks, provide complementary but noisy signals of the underlying 3D scene.
We propose a system that fuses these noisy estimates together into a globally coherent representation of both
the scene geometry and motion. We show a schematic of our pipeline in Figure~\ref{fig:pipeline}.

\subsection{Preliminaries: 3D Gaussian Splatting}
\label{sec:3dgs}
We represent the appearance and geometry of a dynamic scene  with a global set of 3D Gaussians, an explicit and expressive differentiable scene representation~\cite{kerbl20233d} for efficient optimization and rendering.
We define parameters of each 3D Gaussian in the canonical frame $t_0$
as $\vec g_0 \equiv (\gmean_0, \grot_0, \gscale, \gopacity, \gcolor)$, where 
$\gmean_0 \in \mb{R}^3$, $\grot_0 \in \mb{SO}(3)$ are the 3D mean and orientation in the canonical frame, 
and $\gscale \in \mb{R}^3$ the scale, $\gopacity \in \mb{R}$ the opacity, and $\gcolor \in \mb{R}^3$ the color. $\Sigma_0$ is the canonical frame covariance matrix. These properties are persistent and shared across time.
To render a set of 3D Gaussians from a camera with world-to-camera extrinsics $\tf E$ and intrinsics $\tf K$,
the projections of the 3D Gaussians in the image plane are obtained by 2D Gaussians parameterized
as $\gmean'_0 \in \mb{R}^2$ and $\gcovariance'_0 \in \mb{R}^{2\times 2}$ via affine approximation
\begin{align}
    \gmean'_0(\tf K, \tf E) &= \Pi(\tf K \tf E \gmean_0) \in \mb{R}^2, \nonumber \\
    \mb \gcovariance'_0(\tf K, \tf E) &= \tf J_{\tf K \tf E} \tf \Sigma_0 \tf J_{\tf K \tf E}^\top \in \mb{R}^{2\times 2}¥,
\end{align}
where $\Pi$ is perspective projection, and $\tf J_{\tf K \tf E}$ is the Jacobian of perspective projection with $\tf K$ and $\tf E$ at the point $\tf \mu_0$.
The 2D Gaussians can then be efficiently rasterized into RGB image and depth map via alpha compositing as
\begin{equation}
\label{eq:render-3dgs}
\begin{gathered}
    \hat{\mathbf{I}}(\mathbf{p}) = \sum_{i \in H(\mathbf{p})} T_i \alpha_i \gcolor_i,\quad
    \hat{\mathbf{D}}(\mathbf{p}) = \sum_{i \in H(\mathbf{p})} T_i \alpha_i \mathbf{d}_i,\quad
\end{gathered}
\end{equation}
where $\alpha_i = \gopacity_i \cdot \exp{\big(-\frac12 (\mathbf{p} - \gmean'_0)^T\ \gcovariance'_0 \ (\mathbf{p} - \gmean'_0)\big)}$,
and $T_i = \prod_{j=1}^{i-1} (1 - \alpha_j)$.
$H(\mathbf{p})$ is the set of Gaussians that intersect the ray shoot from the pixel $\mathbf{p}$. %
This process is fully differentiable, and enables direct optimization of the 3D Gaussian parameters.

\subsection{Dynamic Scene Representation}
\label{sec:motion_bases}
\noindent \textbf{Scene Motion Parameterization.}
To model a dynamic 3D scene,  we keep track of $N$ canonical 3D Gaussians and vary their positions and orientations over time with a per frame rigid transformation. 
In particular, for a moving 3D Gaussian at time $t$, its pose parameters $(\gmean_t, \tf R_t)$ are rigidly transformed from the canonical frame $t_0$ to $t$ via
$\ftf{\tf T}{t}{0} = 
\begin{bmatrix}
\ftf{\tf R}{t}{0} & \ftf{\tf t}{t}{0}
\end{bmatrix} \in \mb{SE}(3)$: %
\begin{equation}
    \gmean_t = \ftf{\tf R}{t}{0} \tf \mu_0 + \ftf{\tf t}{t}{0},\quad \tf {R}_t = \ftf{\tf R}{t}{0} \tf{R}_0.
\end{equation}
Rather than modeling the 3D motion trajectories independently for each Gaussian,
we define a set of $B \ll N$ learnable basis trajectories $\{\ftf{\tf T}{t}{0}^{(b)} \}_{b=1}^{B}$ that are globally shared across all Gaussians \cite{kratimenos2024dynmf}.
The transformation $\mathbf{T}_{0 \rightarrow t}$ at each time $t$ is then computed by a weighted combination of this global set of basis trajectories through per-point basis coefficients $\tf{w}^{(b)}$ via
\begin{equation}
    \mathbf{T}_{0 \rightarrow t} = \sum_{b=0}^B \motioncoef^{(b)}\ \mathbf{T}_{0 \rightarrow t}^{(b)},
\end{equation}
where $\| \tf{w}^{(b)}\| = 1$.
{In our implementation, we parameterize $\mathbf{T}_{0 \rightarrow t}^{(b)}$ as 6D rotation~\cite{hempel20226d} and translation, and perform the weighted combination separately on each with the same weight $\tf{w}^{(b)}$. 
}
During optimization, we jointly learn the set of global motion bases and motion coefficients of each 3D Gaussian. These compact motion bases explicitly regularize the trajectories to be low-dimensional, encouraging the 3D Gaussians that move similarly to each other to be represented by similar motion coefficients.

\vspace{.5em}
\noindent \textbf{Rasterizing 3D Trajectories.}
Given this representation, we now describe how we obtain pixelwise 3D motion trajectory at any query frame $I_t$. we take a similar approach to 
Wang~\etal~\cite{wang2023omnimotion} and rasterize the motion trajectories of 3D Gaussians into query frame $I_t$.
Namely, for a query camera at time $t$ with intrinsics $\tf K_t$ and extrinsics $\tf E_t$,
we perform rasterization to obtain a map $\world \ftf{\hat{\tf{X}}}{t'}{t} \in \mathcal{R}^{H \times W \times 3}$ that contains
the expected 3D world coordinates of the surface points  corresponding to each pixel at target time $t'$
\begin{equation}
    \world \ftf{\hat{\tf{X}}}{t'}{t}(\mathbf{p}) = \sum_{i \in H(\mathbf{p})} T_i \alpha_i \gmean_{i, t'},
\end{equation}
where $H(\mathbf{p})$ is the set of Gaussians that intersect the pixel $\mathbf{p}$ at query time $t$.
The 2D correspondence location at time $t'$ for a given pixel $\mathbf{p}$, $\hat{\mathbf{U}}_{t \rightarrow t'} (\mathbf{p})$, 
and the corresponding depth value at time $t'$,
$\hat{\mathbf{D}}_{t \rightarrow t'} (\mathbf{p})$
can then be written as
\begin{equation}
    \label{eq:render-2d-track}
    \hat{\tf{U}}_{t \rightarrow t'} (\mathbf{p}) =
    \Pi \big(\tf{K}_{t'} \cam \hat{\tf{X}}_{t \rightarrow t'} (\mathbf{p}) \big),~
    \hat{\mathbf{D}}_{t \rightarrow t'} (\mathbf{p}) = \left(\cam \hat{\tf{X}}_{t \rightarrow t'} (\mathbf{p}) \right)_{[3]},
\end{equation}
where $\cam \hat{\tf{X}}_{t \rightarrow t'} (\mathbf{p}) = \tf{E}_{t'} \world \hat{\tf{X}}_{t \rightarrow t'} (\mathbf{p})$,
$\Pi$ is a perspective projection operation, and $\left( \cdot \right)_{[3]}$ is the third element of a vector.

\subsection{Optimization}
\label{sec:opt}

We prepare the following estimates using off-the-shelf methods in our optimization: 1) camera parameters estimated by MegaSaM~\cite{li2024_MegaSaM}, 2) masks for the moving objects for each frame $\{\mathbf{M}_t\}$, which can be easily obtained using Track-Anything~\cite{Kirillov2023SegmentA,yang2023track} with a few user clicks, 3) monocular depth maps $\{\mathbf{D}_t\}$ computed using state-of-the-art relative depth estimation method such as Depth Anything~\cite{depthanything} and 4) long-range 2D tracks $\{\mathbf{U}_{t \rightarrow t'}\}$ for foreground pixels from state-of-the-art point tracking method TAPIR~\cite{doersch2023tapir}. {We align the relative depth maps with the metric depth maps by computing a per-frame global scale and shift and use them for our optimization, as we found relative depth maps tend to contain finer details}. We treat the lifted 2D tracks unprojected with the aligned depth maps as noisy initial 3D track observations $\{ \tf X_t \}$ for the moving objects. For the static part of the scene, we model them using standard static 3D Gaussians and initialize their 3D locations by unprojecting them into the 3D space using the aligned depth maps. The static and dynamic Gaussians are jointly optimized and rasterized together to form an image. We describe the optimization process %
below.

\vspace{1em}
\noindent \textbf{Initialization.}
We first select the canonical frame $t_0$ to be the frame in which the most 3D tracks are visible,
and initialize the Gaussian means in the canonical frame $\gmean_0$ as $N$ randomly sampled 3D tracks locations from this set of initial observations.
We then perform k-means clustering on the vectorized velocities of the noisy 3D tracks from every frame $\{ {\tf X}_t \}$, and initialize the motion bases $\{ \tf{T}_{0 \rightarrow t}^{(b)} \}_{b=1}^{B}$ from these $B$ clusters of tracks.
Specifically, for the set of trajectories $\{\tf X_t\}_b$ belonging to cluster $b$,
we initialize the basis transform $\ftf{\tf T}{\tau}{0}^{(b)}$ using weighted Procrustes alignment between the point sets $\{\tf X_0\}_b$ 
and $\{\tf X_\tau\}_b$ for all $\tau = 0, \dots, T$, 
where the weights are computed using uncertainty and visibility scores from TAPIR predictions. We initialize $\tf w^{(b)}$ of each Gaussian to decay exponentially with its distance from the center of cluster $b$ in the canonical frame.
We then optimize  $\gmean_0$, $\tf w^{(b)}$, and set of basis functions $\{ \tf{T}_{0 \rightarrow t}^{(b)} \}_{b=1}^{B}$ to fit the observed 3D tracks with an $\ell_1$-loss
under temporal smoothness constraints.

\vspace{.5em}
\noindent \textbf{Training.}
We supervise the dynamic Gaussians with two sets of losses.
The first set comprise our reconstruction loss to match the per-frame pixelwise color, depth, and masks inputs.
The second set enforce consistency of correspondences across time.
Specifically, during each training step, we render the image $\hat{\mathbf{I}}_t$, depth $\hat{\mathbf{D}}_t$, and mask $\hat{\tf M}_t$ from their corresponding training cameras $(\tf K_t, \tf E_t)$ according to Equation~\ref{eq:render-3dgs}.
We supervise these predictions with a reconstruction loss applied independently per-frame
\begin{equation}
    L_\textrm{recon} = \| \hat{\tf I} - \tf I \|_1 + \lambda_\text{depth} \| \hat{\tf D} - \tf D \|_1 + \lambda_\textrm{mask} \| \hat{\tf M} - \tf M \|_1.
\end{equation}

The second set of losses supervises the motion of the Gaussians between frames.
Specifically, we additionally render the 2D tracks $\ftf{\hat{\tf u}}{t'}{t}$ and reprojected depths $\ftf{\hat{\tf D}}{t'}{t}$, for a pair of randomly sampled query time $t$ and target time $t'$. 
We supervise these rendered correspondences with the long-range 2D track estimates via
\begin{align}
    L_{\text{track-2d}} = \| \mathbf{U}_{t \rightarrow t'} - \hat{\mathbf{U}}_{t \rightarrow t'} \|_1, \\
    L_{\text{track-depth}} = \| \hat{\mathbf{d}}_{t \rightarrow t'} - \hat{\mathbf{D}}(\mathbf{U}_{t \rightarrow t'}) \|_1.
\end{align}

Finally, we leverage physics motion prior by enforcing a distance-preserving loss between randomly sampled dynamic Gaussians and their k-nearest neighbors. Let $\hat{\mathbf{X}}_t$ and $\hat{\mathbf{X}}_{t'}$ denote the location of a Gaussian at time $t$ and $t'$, and $\mc{C}_k(\hat{\mathbf{X}}_t)$ denote the set of k-nearest neighbors of  $\hat{\mathbf{X}}_t$, we define
\begin{equation}
    L_{\textrm{rigidity}} = \Big\|\text{dist}(\hat{\mathbf{X}}_t, \mc{C}_k(\hat{\mathbf{X}}_t))
    -
    \text{dist}(\hat{\mathbf{X}}_{t'}, \mc{C}_k(\hat{\mathbf{X}}_{t'}))
    \Big\|_2^2,
\end{equation}
where $\text{dist}(\cdot, \cdot)$ measures Euclidean distance.

\vspace{.5em}
\noindent \textbf{Implementation Details.}
For in-the-wild videos, MegaSaM~\cite{li2024_MegaSaM} is used to estimate initial camera poses, which are further refined as learnable parameters during our optimization, updated via the camera gradient function implemented in the gsplat~\cite{ye2024gsplatopensourcelibrarygaussian} CUDA kernel. For evaluation on public benchmarks, we use the camera annotations provided with each dataset (e.g., from COLMAP~\cite{schonberger2016structure} or simulation).

We optimize our model using Adam Optimizer~\cite{kingma2020method}. We perform $1000$ iterations of optimization for the initialization phase and $500$ epochs for training phase, respectively. The number of $\SETHREE$ bases $B$ is set to 10 for all of our experiments. We initialize $40$k dynamic Gaussians for the dynamic part and $100$k static Gaussians for the static part of the scene, respectively. We perform the same adaptive Gaussian controls for dynamic and static Gaussians as per 3D-GS~\cite{kerbl3Dgaussians}. {Training on a sequence of $300$ frames of $960 \times 720$ resolution takes about 2 hours to finish on an A100 GPU.} Our rendering FPS is around 140 fps. We use the same set of hyper-parameters for all experiments.

\section{Experiment}
\label{sec:experiment}
We evaluate our performance quantitatively and qualitatively on a broad range of tasks: long range 3D point tracking, long-range 2D point tracking, and novel view synthesis. We focus our evaluation on datasets that exhibit substantial scene motion.
In particular, the iPhone dataset~\cite{gao2022dynamic} 
features casual captures of real-world scenes that closely match our target scenarios. It provides comprehensive annotations, including simultaneous validation views, lidar depth, sparse 2D point correspondences across the entire video, and can be used to evaluate our performance on all three tasks. Given the challenge of obtaining precise 3D track annotations for real data, we also evaluate performance using the scenes from the synthetic MOVi-F Kubric dataset~\cite{greff2022kubric}.

\begin{table*}[t]
\centering
\setlength{\tabcolsep}{0.25em} %
\resizebox{0.65\linewidth}{!}{ 
\begin{tabular}{l|ccc|ccc|ccc}
\toprule
\multirow{2}{*}{\textbf{Method}} & \multicolumn{3}{c|}{\textbf{3D Tracking}} & \multicolumn{3}{c|}{\textbf{2D Tracking}} & \multicolumn{3}{c}{\textbf{View Synthesis}} \\
& EPE $\downarrow$ & $\deltathreedsmall\uparrow$ & $\deltathreedlarge\uparrow$ & $\text{AJ}\uparrow$ & $\deltaavg\uparrow$ & $\text{OA}\uparrow$ & PSNR $\uparrow$ & SSIM $\uparrow$ & LPIPS $\downarrow$\\
\midrule
T-NeRF~\cite{gao2022dynamic} & - & - & - & - & - & - & 15.60 & 0.55 & 0.55 \\
HyperNeRF~\cite{park2021hypernerf} & 0.182 & 28.4 & 45.8  & 10.1 & 19.3 & 52.0 & 15.99 & 0.59 & 0.51 \\
DynIBaR~\cite{li2023dynibar} & 0.252 & 11.4 & 24.6  & 5.4 & 8.7 & 37.7 & 13.41 & 0.48 & 0.55 \\
Deformable-3D-GS~\cite{yang2023deformable3dgs} & 0.151 & 33.4 & 55.3  & 14.0 & 20.9 & 63.9 & 11.92 & 0.49 & 0.66 \\
DynMF~\cite{kratimenos2024dynmf} & 0.188 & 22.9 & 53.8 & 5.5 & 9.5 & 60.5 & 16.54 & 0.59 & 0.49 \\
CoTracker~\cite{karaev2023cotracker}+DA~\cite{depthanything} & 0.202 & 34.3 & 57.9  & 24.1 & 33.9 & 73.0 & - & - & -\\
TAPIR~\cite{doersch2023tapir}+DA~\cite{depthanything} & 0.114 & 38.1 & 63.2  & 27.8 & 41.5 & 67.4 & - & - & - \\
{DELTA (world)~\cite{ngo2025deltadenseefficientlongrange}} & 0.159 & 32.5 & 55.3 & 24.7 & 34.1 & 68.9 & - & - & - \\
{SpatialTracker (world)~\cite{xiao2024spatialtrackertracking2dpixels}} & 0.125 & 37.7 & 63.9 & 24.9 & 36.9 & 73.5 & - & - & - \\
Ours & \textbf{0.082} & 43.0 & \textbf{73.3}  & 34.4 & \textbf{47.0} & 86.6 & 16.72 & 0.63 & 0.45 \\
Ours + 2DGS\cite{Huang2DGS2024} & 0.097 & \textbf{47.3} & 71.3 & \textbf{35.8} & \textbf{47.0} & \textbf{87.3} & \textbf{16.75} & \textbf{0.65} & \textbf{0.40} \\
\bottomrule
\end{tabular}%
}
\captionof{table}{\textbf{Evaluation on iPhone dataset}. Our method achieves SOTA performance all tasks of 3D point tracking, 2D point tracking, and novel view synthesis. The baselines that perform best on 2D and 3D tracking (TAPIR~\cite{doersch2023tapir}+DA~\cite{depthanything}, CoTracker~\cite{karaev2023cotracker}+DA~\cite{depthanything}, DELTA~\cite{ngo2025deltadenseefficientlongrange}, SpatialTracker~\cite{xiao2024spatialtrackertracking2dpixels})
are unable to synthesize novel views of the scene, while the methods that perform best in novel view synthesis %
struggle with or fail to produce 2D and 3D tracks. Our method achieves a significant boost in all three tasks above baselines. We include training details about ``Ours + 2DGS~\cite{Huang2DGS2024}'' in the supplement.}
\label{tab: iphone}
\vspace{-1em}
\end{table*}

\subsection{Task Specification}
\noindent \textbf{{Long-Range 3D Tracking.}}
Our primary task is estimating 3D scene motion for any pixel over the entire video. For this purpose, we extend the metrics for scene flow evaluation introduced in RAFT-3D~\cite{teed2021raft3d} to evaluate long-range 3D tracking. Specifically, we report the 3D end-point-error~(EPE), which measures the Euclidean distance between the ground truth 3D tracks and predicted tracks at each target time step.
In addition, we report the percentage of points that fall within a given threshold of the ground truth 3D location $\deltathreedsmall=5$cm and $\deltathreedlarge=10$cm in metric scale.

\vspace{.5em}
\noindent \textbf{{Long-Range 2D Tracking.}}  Our 3D motion representation can be easily projected onto image plane to get long-range 2D tracks. Thus we also evaluate 2D tracking performance in terms of both position and occlusion accuracy. Following the metrics introduced in the TAP-Vid benchmark~\cite{doersch2022tap}, we report the Average Jaccard~(AJ), average position accuracy~($\deltaavg$), and Occlusion Accuracy (OA). 

\vspace{.5em}
\noindent \textbf{{Novel View Synthesis.} }
We measure our method's novel view synthesis  quality as a comprehensive assessment for geometry, appearance, and motion. 
We evaluate on the iPhone dataset~\cite{gao2022dynamic} which provides validation views and report co-visibility masked image metrics~\cite{gao2022dynamic}: mPSNR, mSSIM~\cite{wang2004image} and mLPIPS~\cite{gatys2017controlling,huh2020transforming}. Additionally, we evaluate on NVIDIA dataset for more comparison.

\providecommand\animage{}
\renewcommand{\animage}[2]{
    \frame{\includegraphics[width=\linewidth,clip,trim=#1]{figures/comparisons_iphone_tracks_3d/#2}}
}
\begin{figure}[t!]
    \setlength{\tabcolsep}{0.8pt}
    \renewcommand{\arraystretch}{0.5}
    \begin{tabularx}{0.48\textwidth}{@{}c*{4}{C}@{}}
        \makebox[20pt]{\raisebox{30pt}{\rotatebox[origin=c]{90}{\footnotesize \textsc{Backpack}}}}\hspace{-4pt} &
        \animage{0 120 0 0}{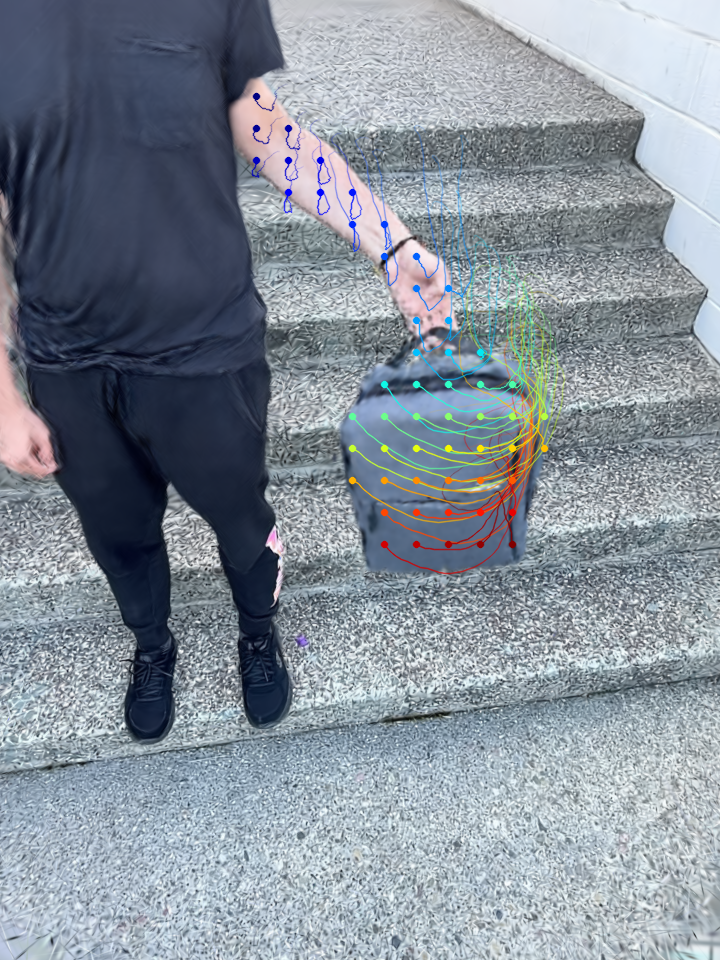} &
        \animage{0 120 0 0}{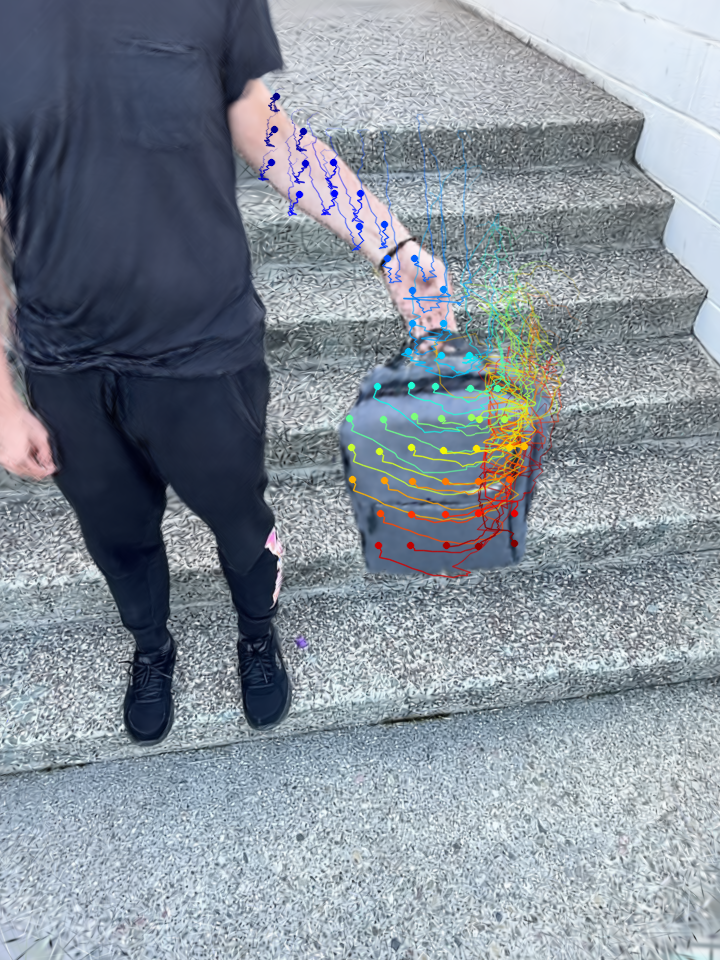} &
        \animage{0 120 0 0}{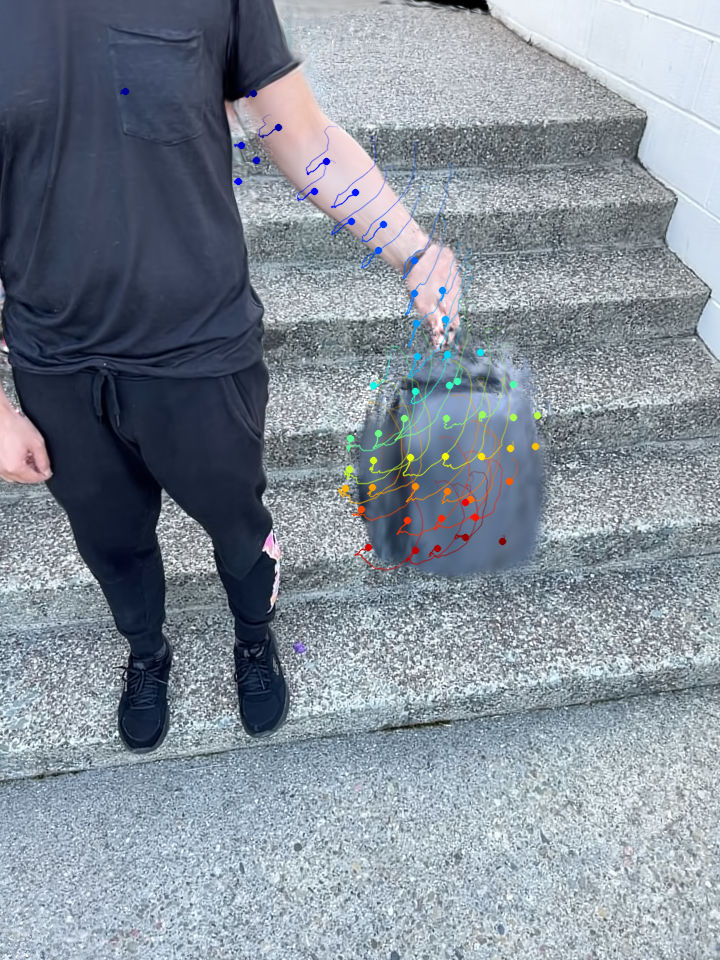} &
        \animage{0 120 0 0}{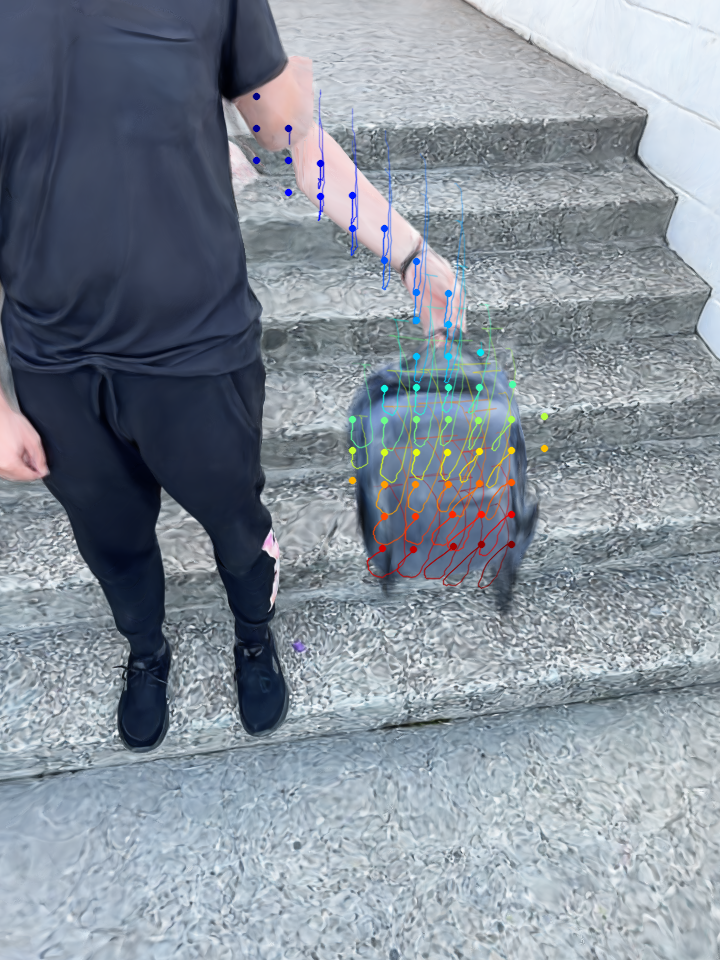}
        \\
        \makebox[20pt]{\raisebox{30pt}{\rotatebox[origin=c]{90}{\footnotesize \textsc{Paper-windmill}}}}\hspace{-4pt} &
        \animage{0 60 0 60}{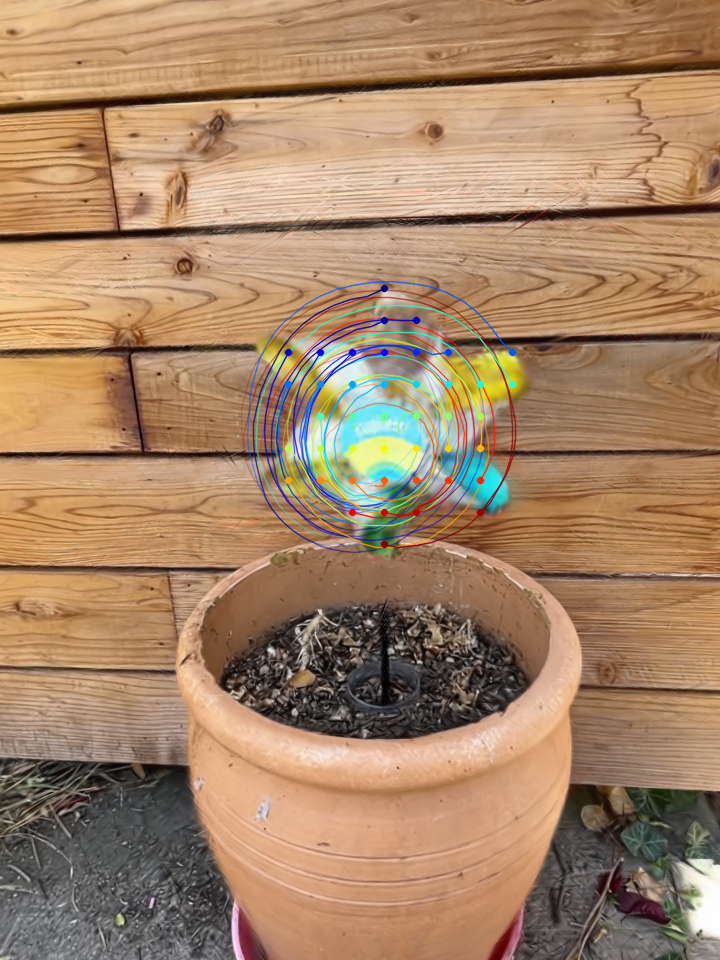} &
        \animage{0 60 0 60}{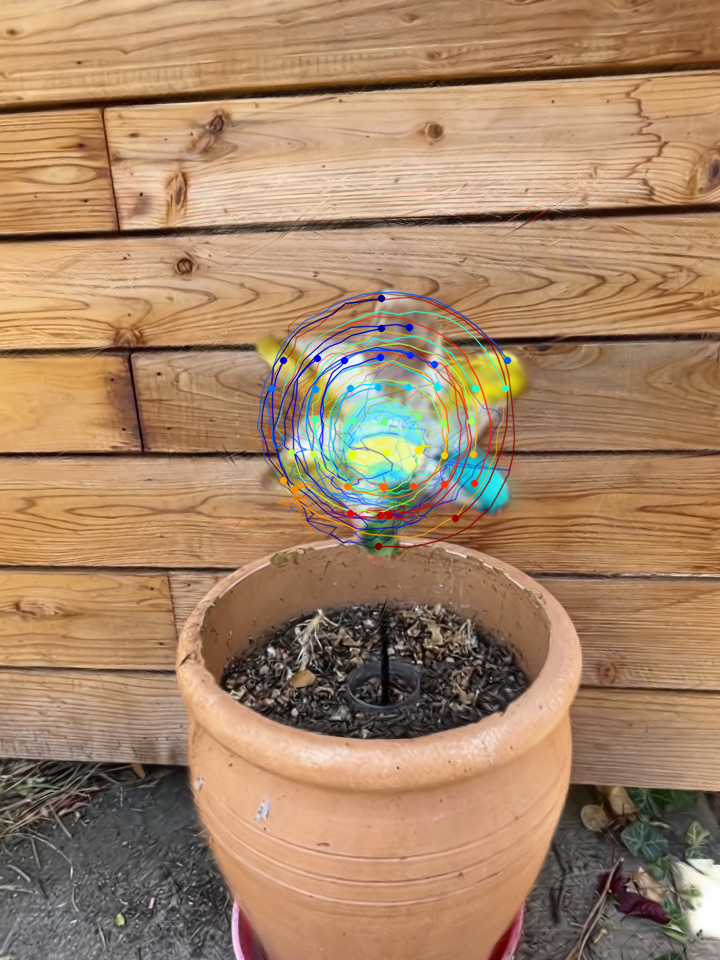} &
        \animage{0 60 0 60}{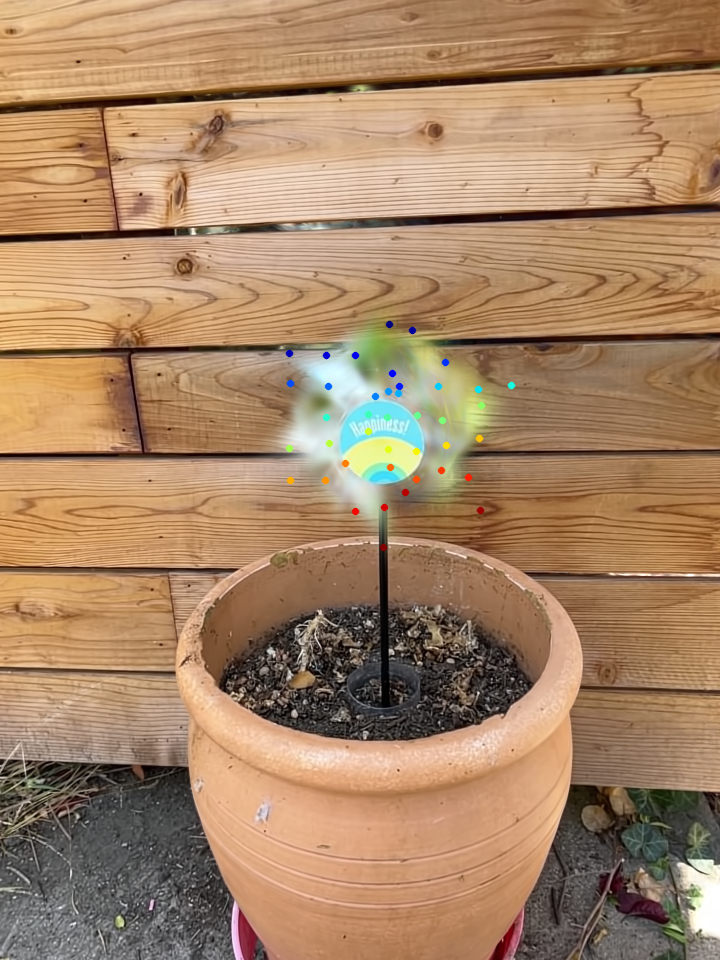} &
        \animage{0 60 0 60}{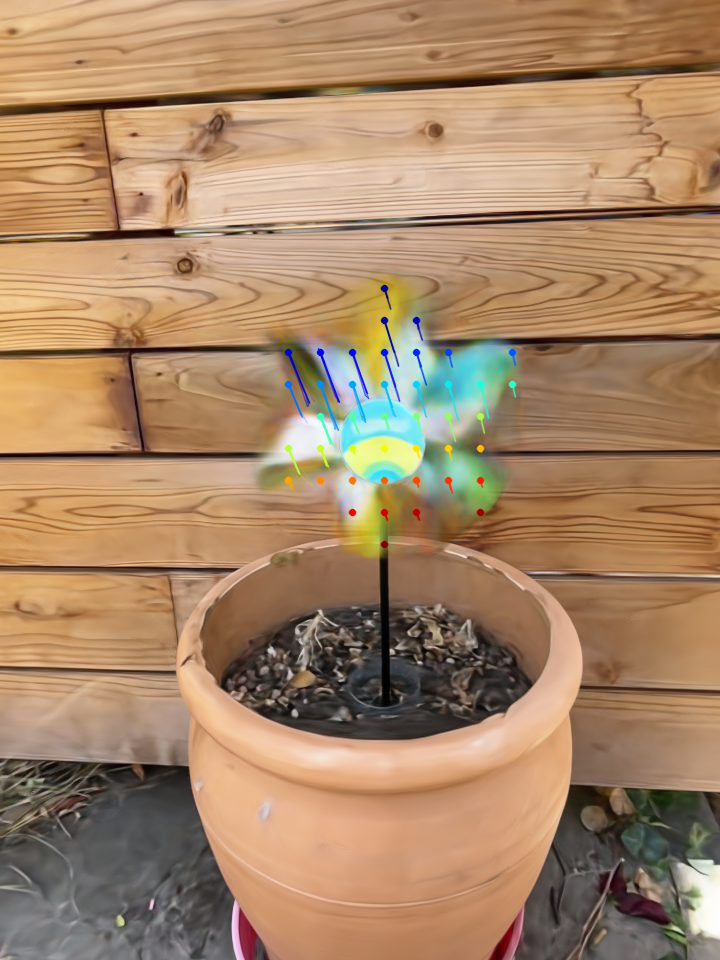}
        \\
        \makebox[20pt]{\raisebox{30pt}{\rotatebox[origin=c]{90}{\footnotesize \textsc{Spin}}}}\hspace{-4pt} &
        \animage{0 60 0 60}{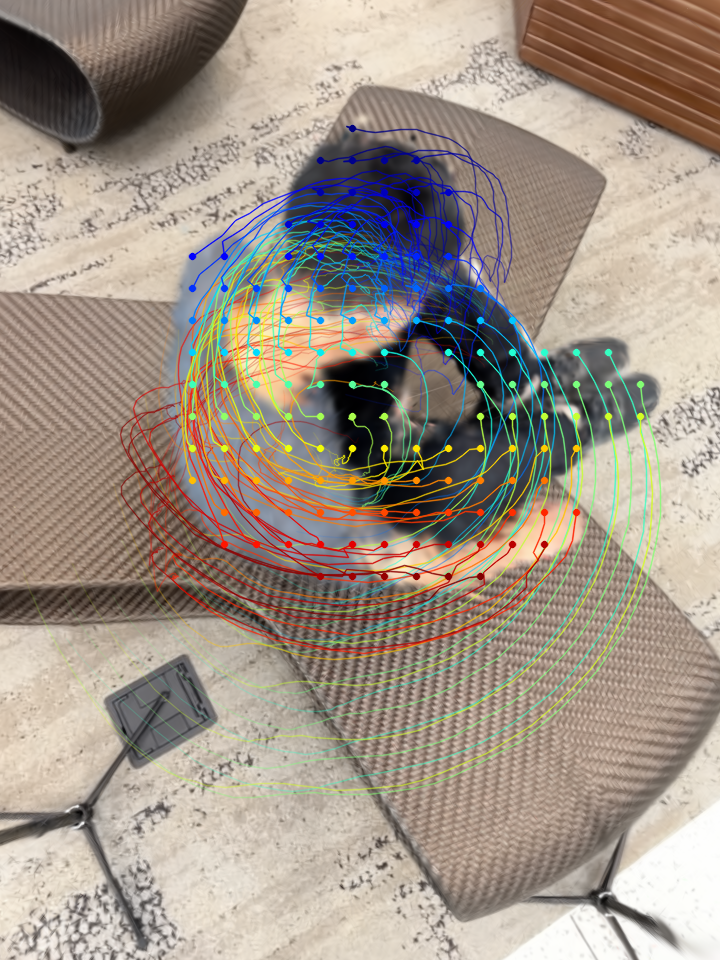} &
        \animage{0 60 0 60}{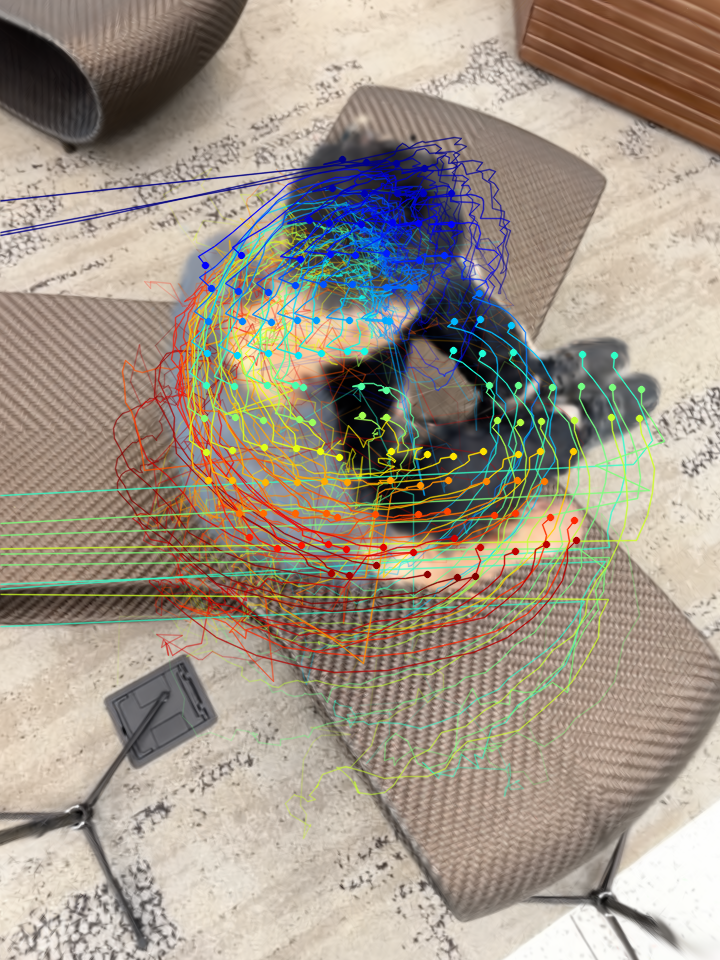} &
        \animage{0 60 0 60}{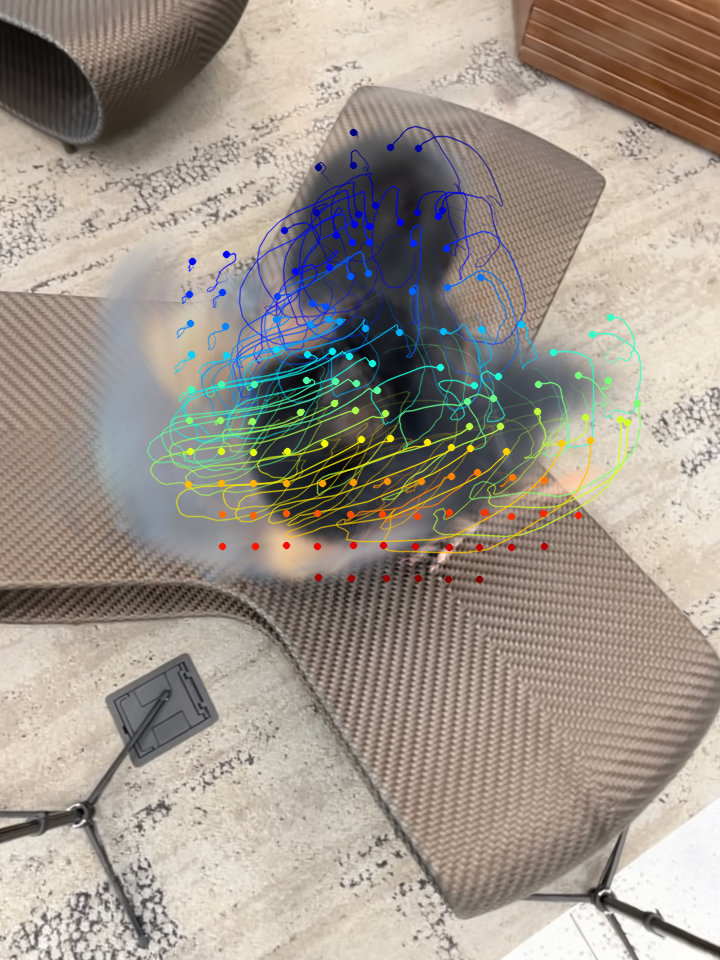} &
        \animage{0 60 0 60}{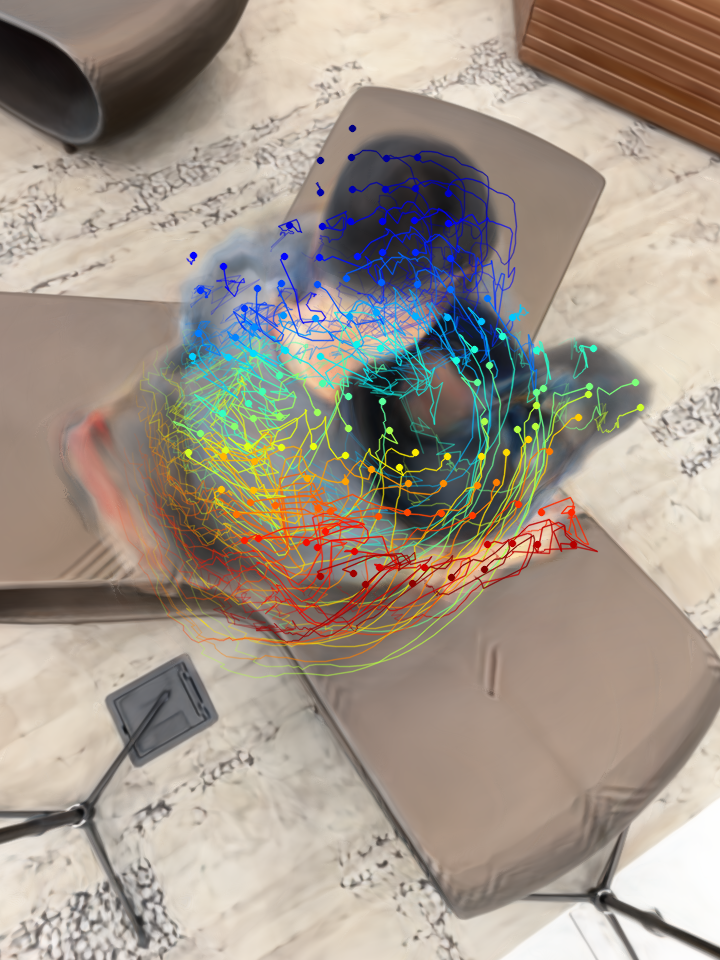}
        \\
        [2pt] &
        {\small Ours} &
        {\small TAPIR + DA$^*$} &
        {\small D-3DGS} &
        {\small HyperNeRF}
    \end{tabularx}
    \captionof{figure}{
        \textbf{3D Track visualization on iPhone dataset.} We render novel views for each method and overlay their predicted 3D tracks on top of the rendered images. For clarity, we only display a segment of the trails spanning 50 frames for a specific set of grid query points. However, it is important to note that our method can generate dense, full-length 3D tracks. 
        {$^*$Note that TAPIR + DA cannot produce novel view rendering, and we overlay their tracking results with our rendering to make it easier to interpret.}
    }
    \vspace{-2em}
    \label{fig: 3d_tracking_comparison}
\end{figure}

\providecommand\animage{}
\renewcommand{\animage}[2]{
    \frame{\includegraphics[width=\linewidth,clip,trim=#1]{figures/paper-graphics2/#2}}
}
\begin{figure*}[t!]
    \setlength{\tabcolsep}{0.8pt}
    \renewcommand{\arraystretch}{0.5}
    \begin{tabularx}{\textwidth}{@{}c*{7}{C}@{}}
        \makebox[20pt]{\raisebox{60pt}{\rotatebox[origin=c]{90}{\textsc{\hspace{-40pt} Paper-windmill}}}}\hspace{-6pt} &
        \animage{0 0 0 0}{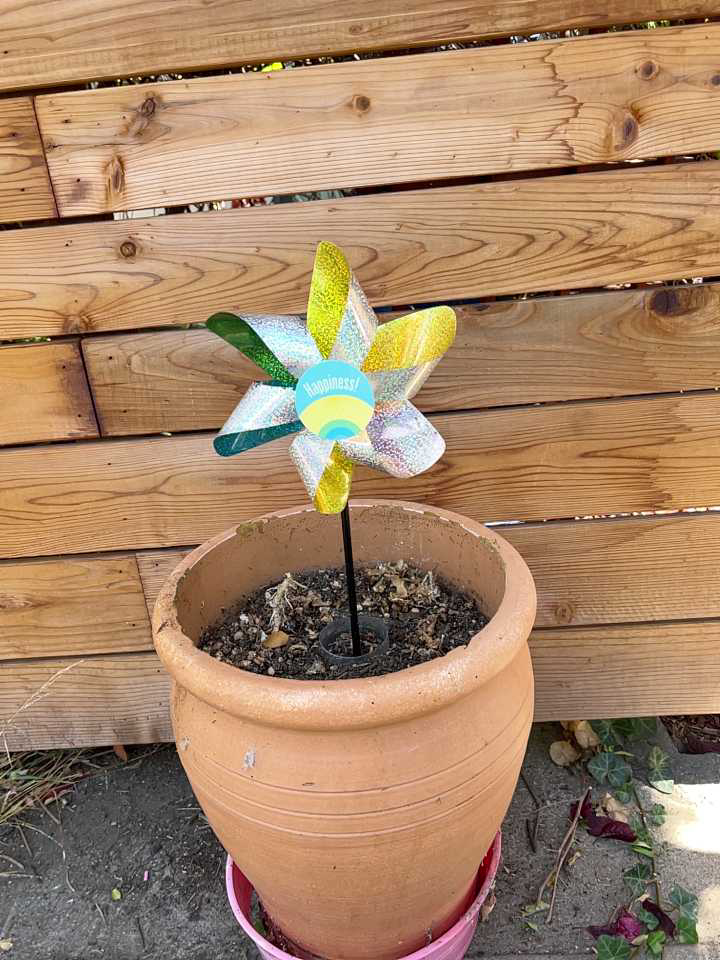} &
        \animage{0 0 0 0}{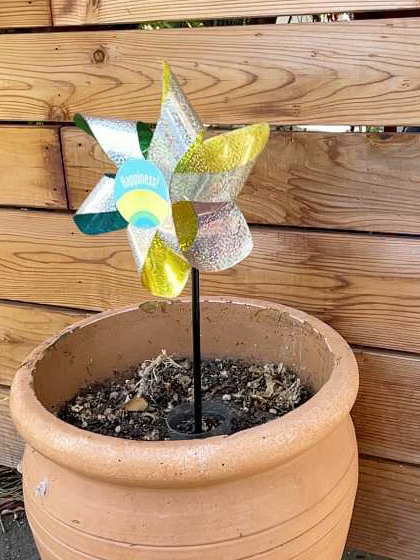} &
        \animage{0 0 0 0}{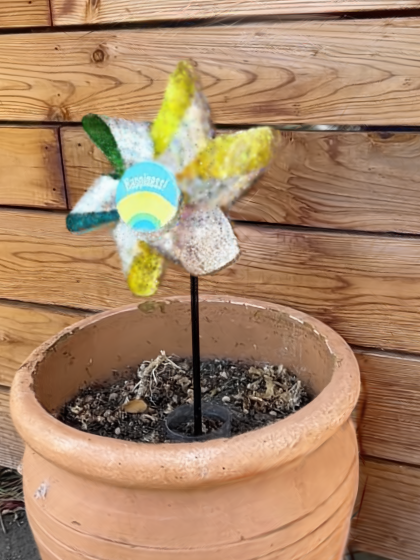} &
        \animage{0 0 0 0}{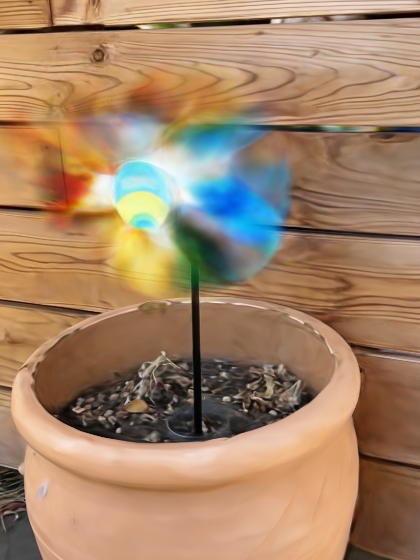} &
        \animage{0 0 0 0}{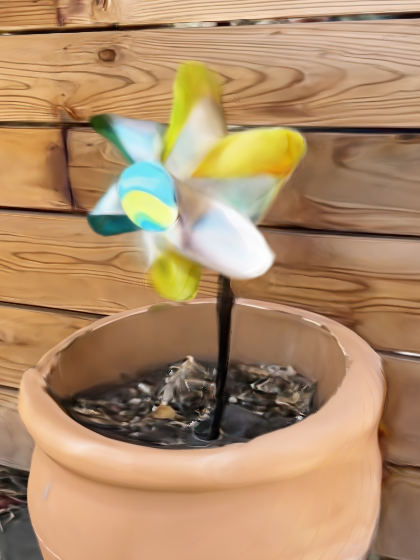} &
        \animage{0 0 0 0}{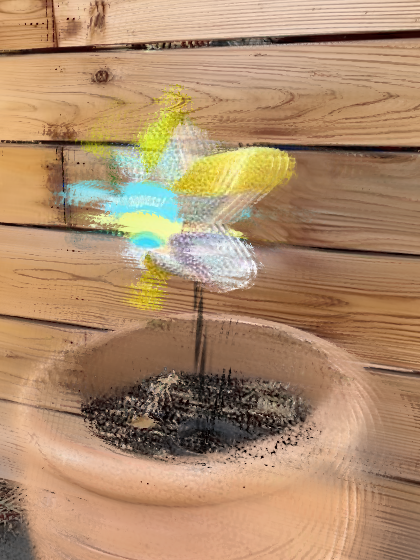} &
        \animage{0 0 0 0}{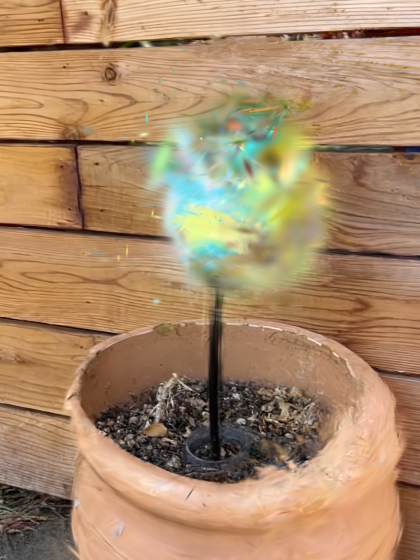}
        \\
        \makebox[20pt]{\raisebox{55pt}{\rotatebox[origin=c]{90}{\textsc{\hspace{-36pt} Spin}}}}\hspace{-6pt} &
        \animage{0 0 0 0}{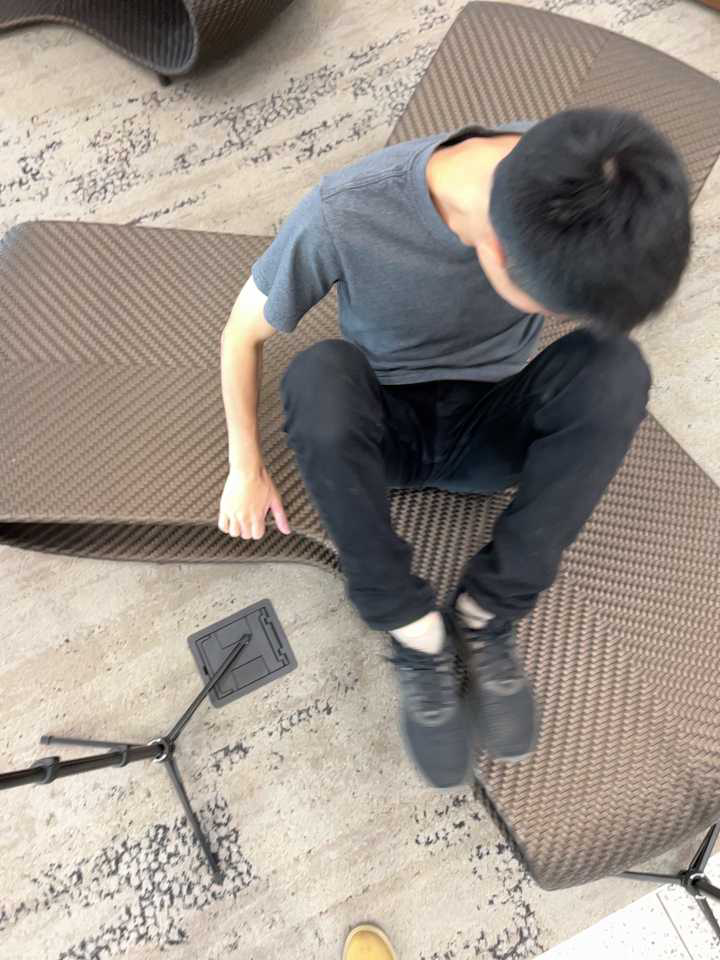} &
        \animage{0 0 0 0}{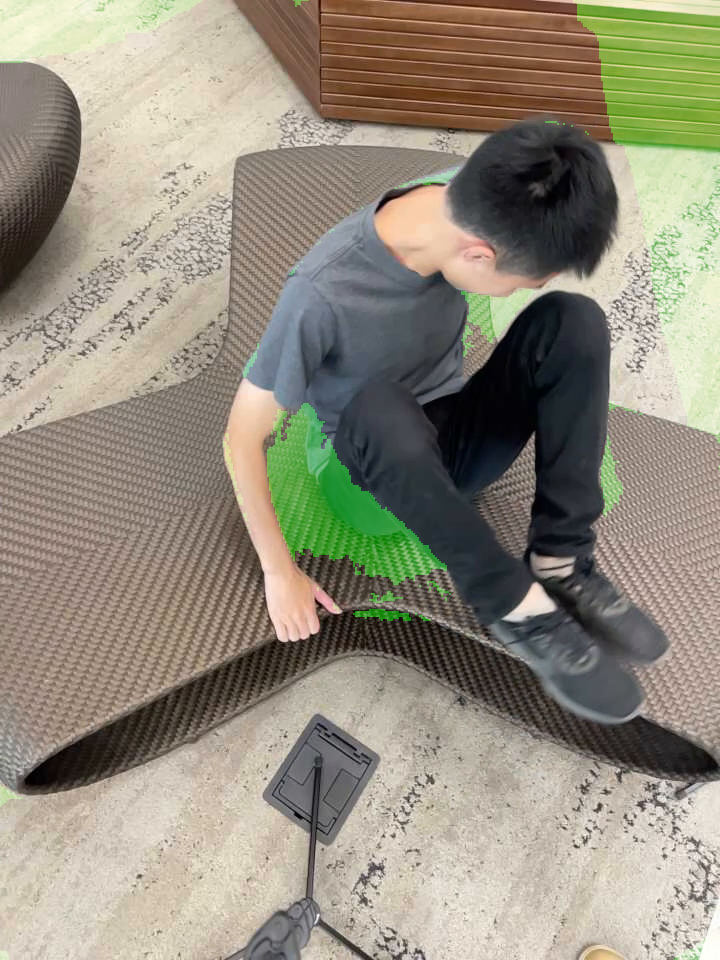} &
        \animage{0 0 0 0}{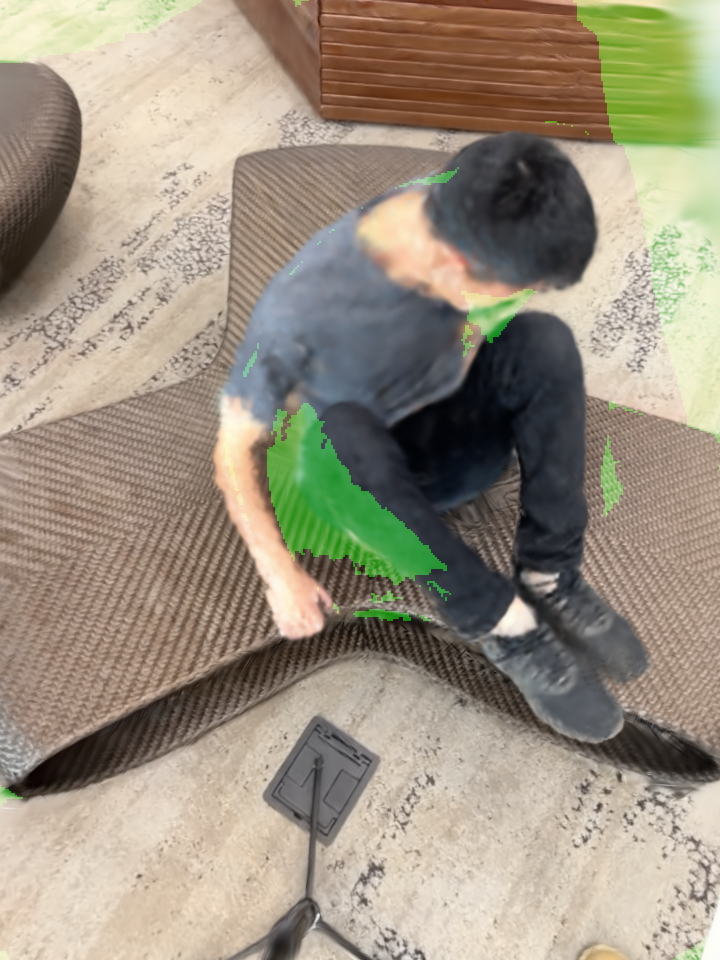} &
        \animage{0 0 0 0}{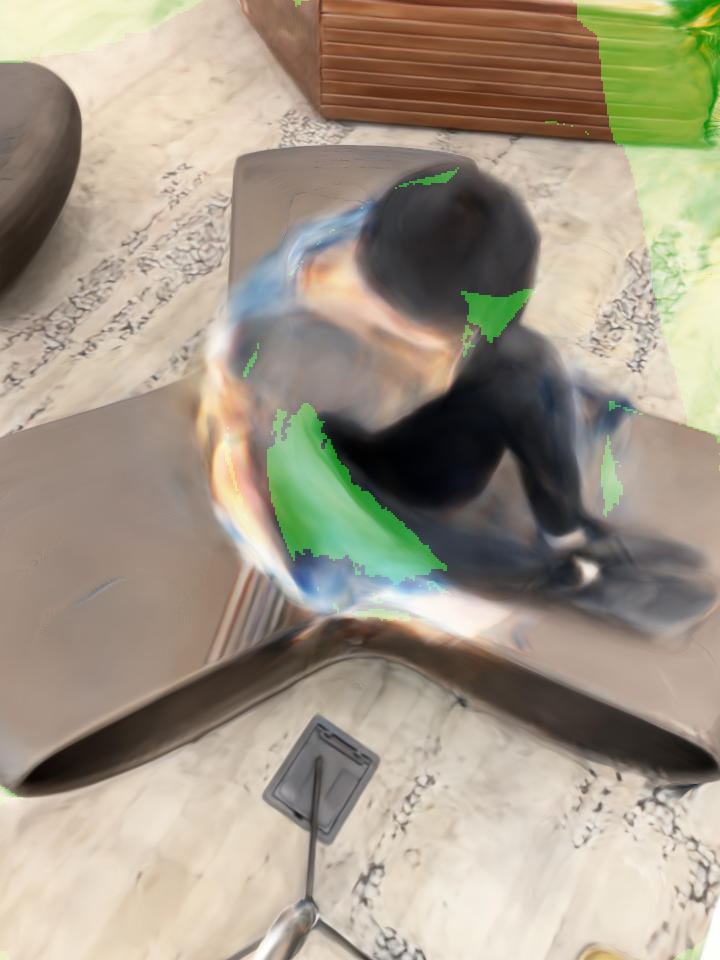} &
        \animage{0 0 0 0}{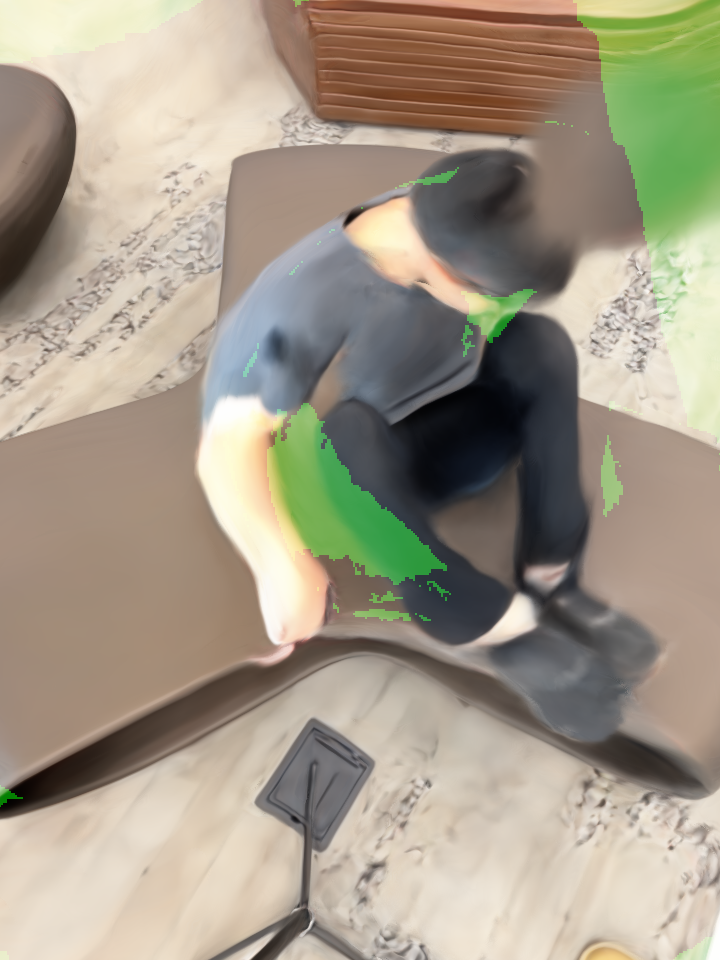} &
        \animage{0 0 0 0}{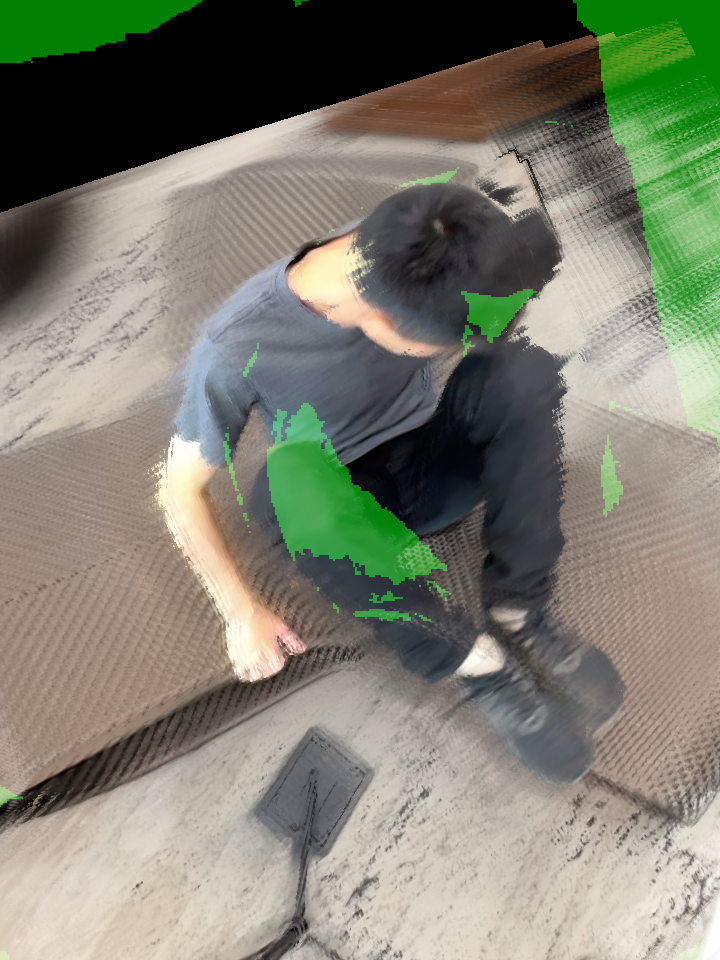} &
        \animage{0 0 0 0}{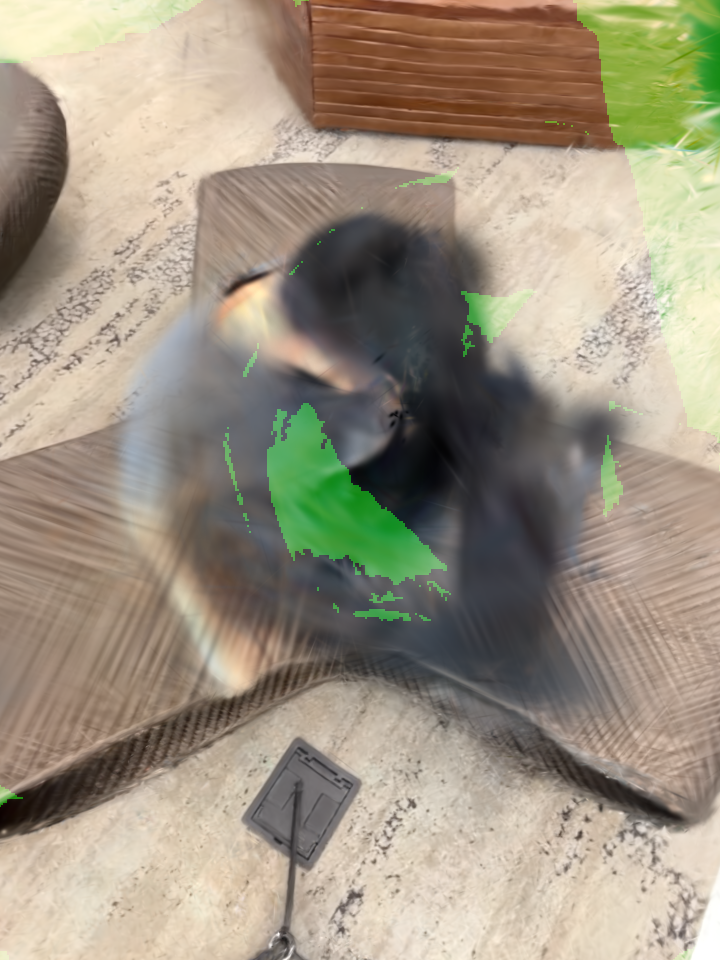}
        \\
        \makebox[20pt]{\raisebox{45pt}{\rotatebox[origin=c]{90}{\textsc{\hspace{-18pt} Teddy}}}}\hspace{-6pt} &
        \animage{0 0 0 0}{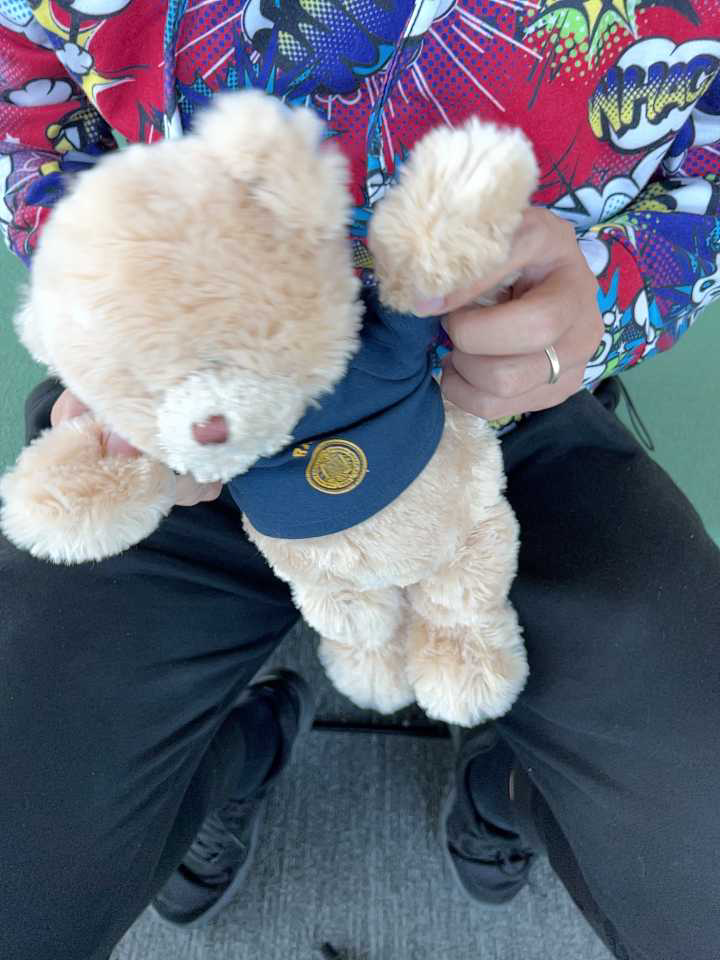} &
        \animage{0 0 0 0}{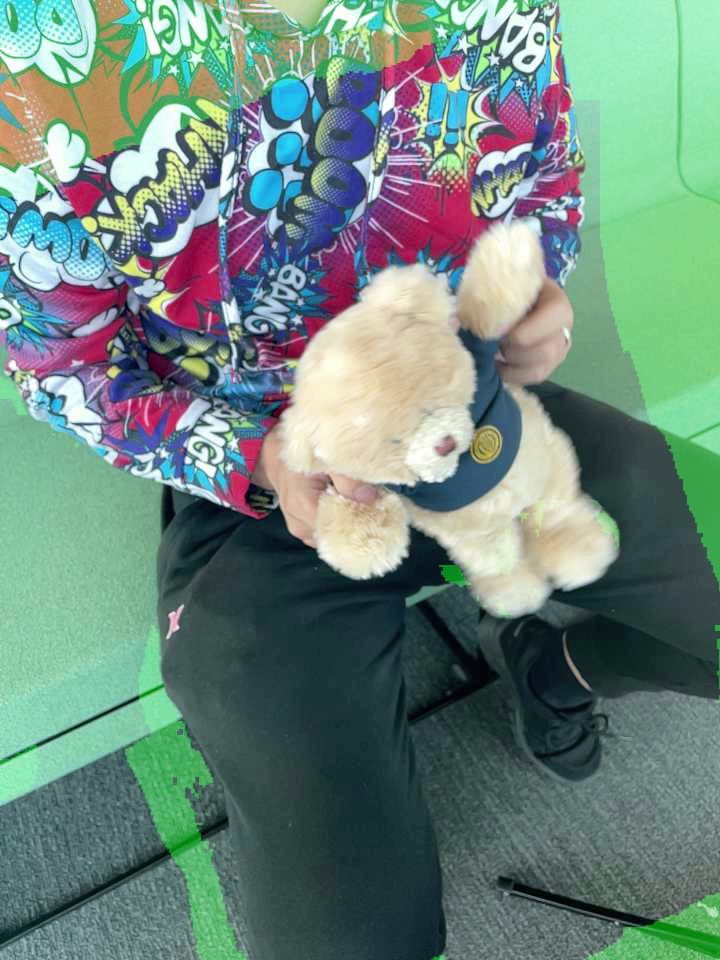} &
        \animage{0 0 0 0}{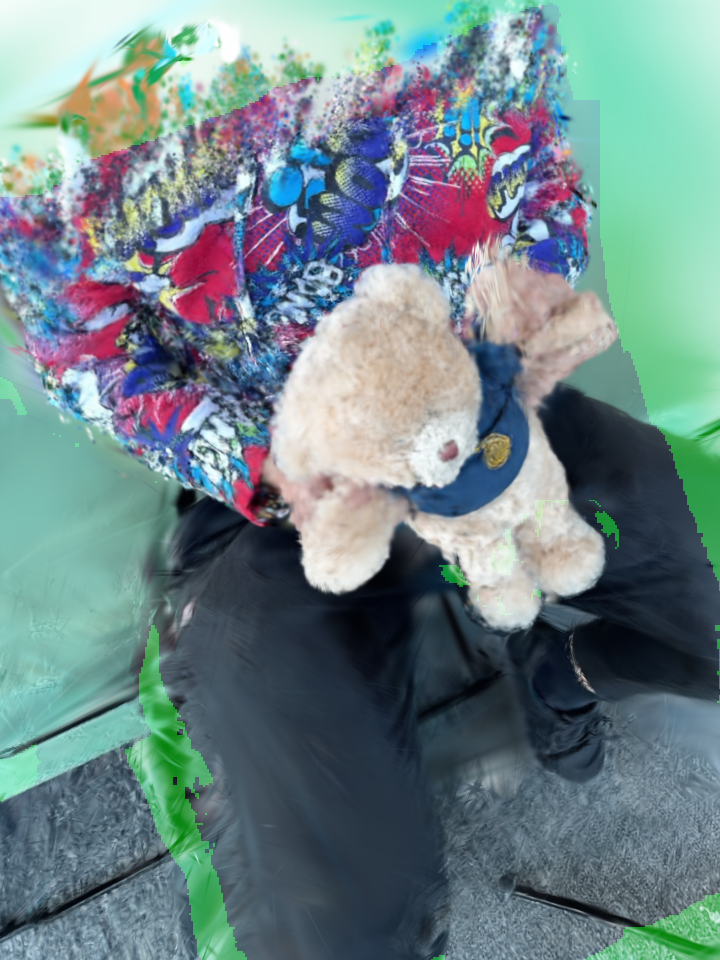} &
        \animage{0 0 0 0}{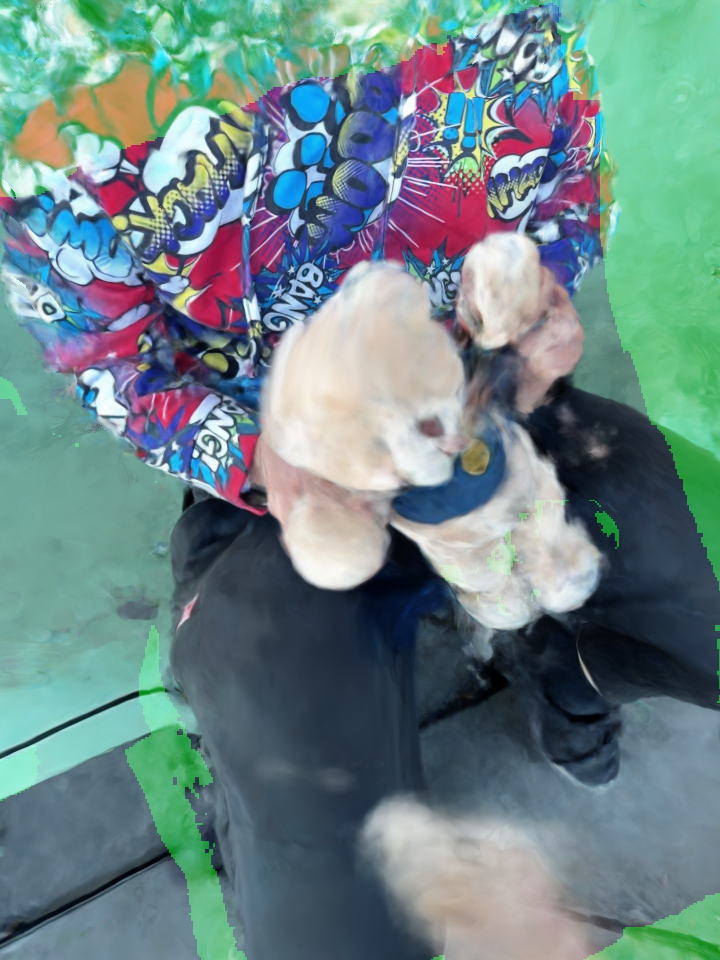} &
        \animage{0 0 0 0}{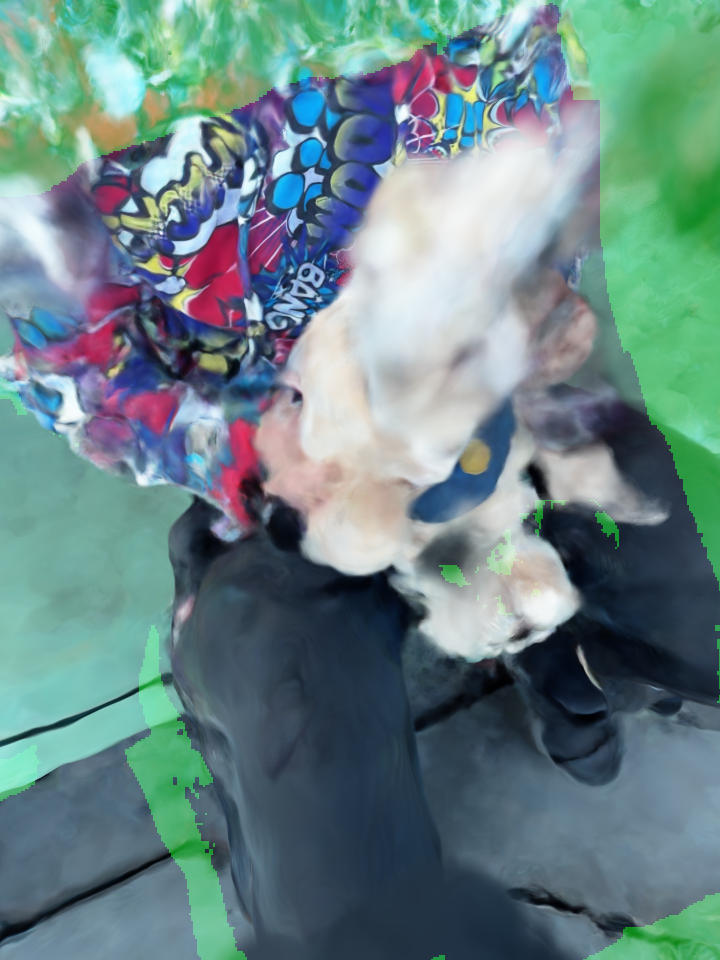} &
        \animage{0 0 0 0}{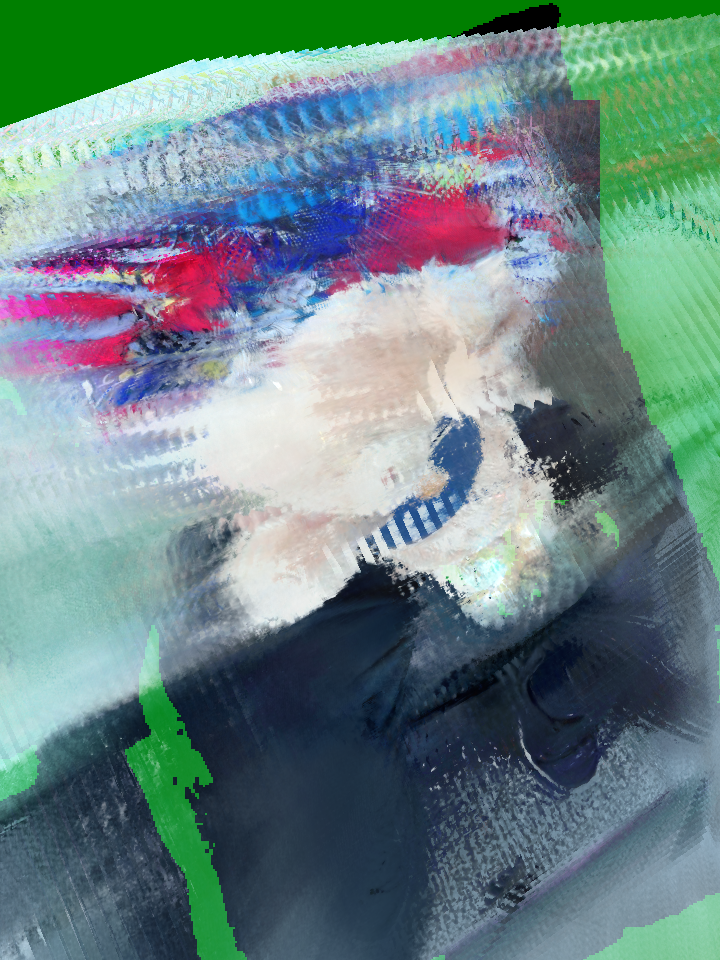} &
        \animage{0 0 0 0}{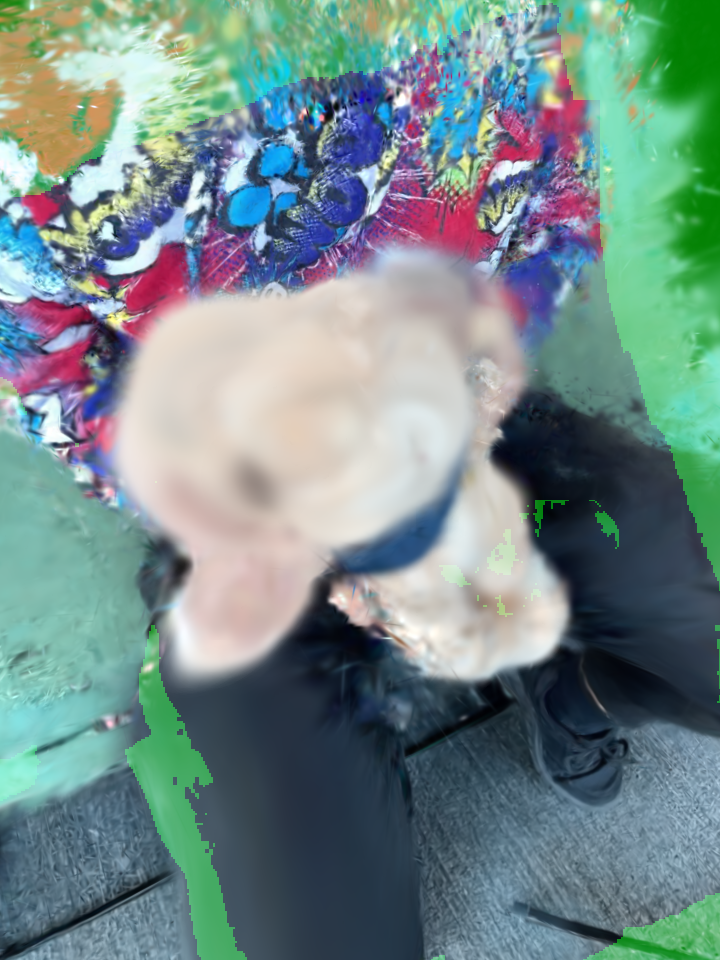}
        \\
        [2pt] &
        {\footnotesize Train View} &
        {\footnotesize GT} &
        {\footnotesize Ours} &
        {\footnotesize  HyperNeRF } &
        {\footnotesize TNeRF } &
        {\footnotesize DynIBaR  } &
        {\footnotesize D-3DGS  }
    \end{tabularx}
    \vspace{-4pt}
    \caption{
        \textbf{Qualitative comparison of novel view synthesis on iPhone dataset.} The leftmost image in each row shows the training view at the same time step as the validation view. The regions highlighted in green indicate areas excluded from evaluation due to the lack of co-visibility between training and validation views. Please see the supplemental video for more qualitative results and comparisons.
    }
    \vspace{-1em}
    \label{fig: novel view comparison}
\end{figure*}

\subsection{Baselines}
Our method represents dynamic 3D scene comprehensively with explicit long-range 3D scene motion estimation, which also allows for novel view synthesis.  
While no existing method achieves the exact same goals as ours, there are methods that focus on sub-tasks related to our problem, such as dynamic novel view synthesis, 2D tracking, or monocular depth estimation.
We therefore adapt existing methods as our baselines which are introduced below.

While dynamic novel view synthesis approaches focus primarily on the photometric reconstruction quality and do not explicitly output 3D point tracks, we can adapt some of their representations to derive 3D point tracks for our evaluation.
For HyperNeRF~\cite{park2021hypernerf}, we compose the learned inverse mapping~(from view space to canonical space) and a forward mapping solved via root-finding~\cite{gao2022dynamic,chen2021snarf} to produce 3D tracks at query points.
DynIBaR~\cite{li2023dynibar} produces short-range view-to-view scene flow, which we chain into long-range 3D tracks for our evaluation.
Deformable-3D-GS~(D-3DGS)~\cite{yang2023deformable3dgs} represents dynamic scenes using 3D Gaussians~\cite{kerbl20233d} in the canonical space and a deformation MLP network that deforms the canonical 3D Gaussians into each view, which naturally allows 3D motion computation. T-NeRF~\cite{gao2022dynamic} models dynamic scenes using time as MLP input in addition to 3D locations, which does not provide a method for extracting 3D motion, and hence they are not considered for 2D/3D tracking evaluation. Finally, we implemented our own version of DynMF~\cite{kratimenos2024dynmf} with neural network motion bases. Implementation details are provided in the supplemental material. 
We evaluate the 3D tracks of these methods only on the iPhone dataset, because we found none can handle the Kubric scene motion.

Alongside dynamic view synthesis baselines, we include 3D tracking baselines. 
Specifically, we take state-of-the-art long-range 2D tracks from TAPIR~\cite{doersch2023tapir} and CoTracker~\cite{karaev2023cotracker} and lift them into 3D scene motion using monocular depth maps produced by Depth Anything~(DA)~\cite{depthanything}. 
We compute the correct scale and shift for each relative depth map from Depth Anything to align them with the scene. 
The two resulting baselines are called ``CoTracker~\cite{karaev2023cotracker} + DA~\cite{depthanything}'' and ``TAPIR~\cite{doersch2023tapir} + DA~\cite{depthanything}''.
Note that these baselines can only produce 3D tracks for visible regions, as the depth values for occluded points are unknown from such a 2.5D representation. 
In contrast, our global representation allows for modeling 3D motion through occlusions. In addition, we include frame-space 3D tracking methods such as DELTA~\cite{ngo2025deltadenseefficientlongrange} and SpatialTracker~\cite{xiao2024spatialtrackertracking2dpixels}, and transform their trajectories into world coordinates using annotated camera poses for evaluation.

\providecommand\animage{}
\renewcommand{\animage}[2]{
    \frame{\includegraphics[width=\linewidth,clip,trim=#1]{figures/qualitative_depth_coefs/#2}}
}
\begin{figure}[t!]
    \setlength{\tabcolsep}{0.8pt}
    \renewcommand{\arraystretch}{0.5}
    \begin{tabularx}{0.47\textwidth}{@{}*{3}{C}*{3}{C}@{}}
        \animage{0 0 1 0}{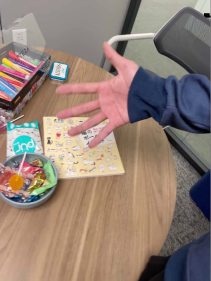} &
        \animage{0 0 0 0}{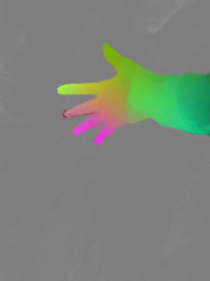} &
        \animage{4 8 5 0}{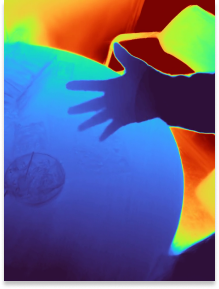} &
        \animage{0 0 1 0}{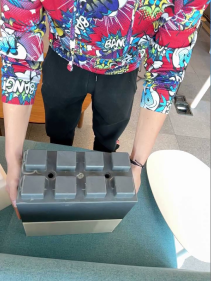} &
        \animage{0 0 0 0}{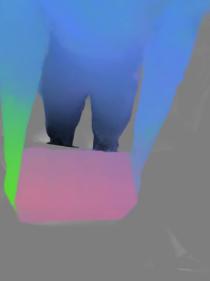} &
        \animage{0 0 1 0}{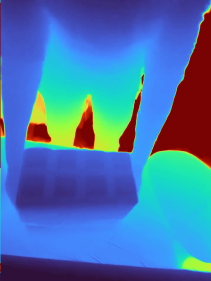} 
        \\
        {\small Input} &
        {\small PCA Coef.} &
        {\small Pred. Depth} &
        {\small Input} &
        {\small PCA Coef.} &
        {\small Pred. Depth}
    \end{tabularx}
    \caption{
        \textbf{Visualization of motion coefficients after PCA and predicted depth maps on iPhone dataset.} The motion coefficients encode information regarding rigid moving components. For instance, our motion coefficient PCA produces constant color for the block in the second example which exhibits rigid motion.
    }
    \label{fig: motion_coefs}
    \vspace{-1em}
\end{figure}
\subsection{Evaluation on iPhone Dataset}
\label{sec:iphone_results}
iPhone dataset~\cite{gao2022dynamic} contains 14 sequences of 200-500 frames featuring various types of challenging real world scene motion.
All sequences are recorded using a handheld moving camera in a casual manner, with 7 of them additionally featuring two synchronized static cameras with a large baseline for novel view synthesis evaluation. 
For 3D tracking evaluation, we generate the groundtruth 3D tracks by lifting the 2D keypoint annotations into 3D using lidar depth, and masking out points that are occluded or with invalid lidar depth values. All experiments are conducted with the original instead of half resolution as in \cite{gao2022dynamic} given that our method can handle high-res video input.
We also discard scenes \textsc{Space-out} and \textsc{Wheel} due to camera and lidar error.
We find that the original camera annotations from ARKit is not accurate enough, and refine them using global bundle adjustment from COLMAP~\cite{schoenberger2016sfm}. The refined poses are used for all methods.

We report the quantitative comparison in Tab.~\ref{tab: iphone} which shows that our method outperforms all baselines on all tasks by a substantial margin. 
For tracking, we observe significant improvements over naive baselines that combine 2D tracks with depth maps, especially ``TAPIR+DA'', which also serves as input to our system, highlighting the effectiveness of our consolidation process.
The improvement stems from three factors: 1) a low-dimensional $\SETHREE$ motion representation, which allows inferring occluded motion via nearby visible regions; 2) 3D regularizations, such as low acceleration constraints that promote coherent trajectories; 3) the consolidation of all pairwise tracks into a single 4D representation with persistent 3D geometry, which
serves as a structural prior to correct noisy 2D tracks.
Our method also achieves the best novel view synthesis quality across dynamic NeRF and 3D-GS-based baselines. 

Fig.~\ref{fig: 3d_tracking_comparison} shows qualitative comparison of the 3D tracking results. We render the novel views and plot the predicted 3D tracks of the given query points onto the novel views. 
Since ``TAPIR + DA'' cannot perform novel view synthesis, we overlay their track predictions onto our renderings to aid interpretation. D-3DGS~\cite{yang2023deformable3dgs} and HyperNeRF~\cite{park2021hypernerf} fail to capture the significant scene motion in \textsc{paper-windmill} and \textsc{spin}, resulting in structure degradation and blurry rendering. ``TAPIR + DA'' can track large motions, but their 3D tracks tend to be noisy and erroneous. In contrast, our method not only generates high-quality novel views but also the most smooth and accurate 3D tracks.
Fig.~\ref{fig: novel view comparison} provides additional novel view synthesis comparison on the validation views. In Fig.~\ref{fig: motion_coefs}, we visualize the rendering of the first three components of PCA decomposition for the motion coefficients, which correlates well with the rigid groups of the moving objects. 

\subsection{Evaluation on Kubric Dataset}
\label{sec:kubric_results}
\providecommand\animage{}
\renewcommand{\animage}[2]{
    \frame{\includegraphics[width=.9\linewidth,clip,trim=#1]{figures/kubric_pca/#2}}
}
\begin{figure}[htbp]
\begin{minipage}[b]{0.45\textwidth}
    \centering
    \footnotesize
    \begin{tabular}{l|ccc}
        \toprule
        \textbf{Method} & EPE$\downarrow$ & $\deltathreedsmall\uparrow$ & $\deltathreedlarge\uparrow$ \\
        \midrule
        CoTracker~\cite{karaev2023cotracker}+DA~\cite{depthanything}  & 0.19 & 34.4 & 56.5  \\
        TAPIR~\cite{doersch2023tapir}+DA~\cite{depthanything} & 0.20 & 34.0 & 56.2  \\
        Ours & \textbf{0.16} & \textbf{39.8} & \textbf{62.2}  \\
        \bottomrule
    \end{tabular}
    \captionof{table}{\textbf{3D Tracking evaluation on Kubric dataset. }}
    \vspace{-2em}
    \label{tab:kubric}
\end{minipage}
\hspace{2cm} %
\end{figure}

\noindent The Kubric MOVi-F dataset contains short 24-frame videos of scenes of 10-20 objects, rendered with linear camera movement and motion blur.
Multiple rigid objects are tossed onto the scene, at a speed that far exceeds the speed of the moving camera,
making it a similar capture scenario to in-the-wild capture scenarios.
It provides dense comprehensive annotations, including ground truth depth maps, camera parameters, segmentation masks,
and point correspondences that are dense across time.
We use 30 scenes from the MOVi-F validation set to evaluate the accuracy of our long-range 3D point tracks for all points in time.

We demonstrate our method on the synthetic Kubric MOVi-F dataset in long-range 3D point tracking across time.
Of the above baselines, only ``CoTracker~\cite{karaev2023cotracker}+DA~\cite{depthanything}'' and ``TAPIR~\cite{karaev2023cotracker}+DA~\cite{depthanything}'' yield 3D tracks for these scenes.
For all baselines, we provide the ground truth camera intrinsics and extrinsics,
and monocular depth estimates that have been aligned to the ground truth depth map.
We obtain point tracks for all non-background pixels for each method.

We report our quantitative 3D point tracking metrics in Table~\ref{tab:kubric},
and find that across all metrics, our method outperforms the baselines.
Please see the supplement for figures of the optimized motion coefficients, which coherently group moving objects.

\subsection{Evaluation on NVIDIA Dataset}

We conduct experiments on seven scenes from the NVIDIA dataset \cite{yoon2020novel}, following the evaluation protocol of Dynamic Gaussian Marbles \cite{stearns2024dynamic}, which evaluates the task from a single static camera instead of interleaving different camera as discussed in ~\cite{gao2022dynamic}. We adopt the same image resolution, camera parameters and depths maps as input. Please see the supplement for more details. 
For each scene, we compute covisibility masks between each evaluation view and the training view at each time step to exclude unobservable regions during evaluation. Results are reported in Tab. \ref{tab:nvidia}.

\begin{table}[h]
\centering
\setlength{\tabcolsep}{0.25em} %
\resizebox{\linewidth}{!}{%
\begin{tabular}{l|ccc|ccc|ccc|ccc}
\toprule
\multirow{2}{*}{\textbf{Method}} &
  \multicolumn{3}{c|}{Balloon1} &
  \multicolumn{3}{c|}{Balloon2} &
  \multicolumn{3}{c|}{Jumping} &
  \multicolumn{3}{c}{Truck} \\
 & PSNR $\uparrow$ & SSIM $\uparrow$ & LPIPS $\downarrow$ 
 & PSNR $\uparrow$ & SSIM $\uparrow$ & LPIPS $\downarrow$ 
 & PSNR $\uparrow$ & SSIM $\uparrow$ & LPIPS $\downarrow$ 
 & PSNR $\uparrow$ & SSIM $\uparrow$ & LPIPS $\downarrow$ \\
\midrule
DGM \cite{stearns2024dynamic} 
  & 23.83 & \textbf{0.81} & 0.066 
  & 23.62 & 0.81 & 0.088 
  & 20.02 & \textbf{0.73} & \textbf{0.134} 
  & 27.31 & 0.86 & 0.055 \\
Ours 
  & \textbf{23.90} & 0.79 & \textbf{0.058} 
  & \textbf{23.91} & \textbf{0.83} & \textbf{0.074} 
  & \textbf{20.08} & 0.69 & 0.151 
  & \textbf{27.65} & \textbf{0.88} & \textbf{0.048} \\
\midrule
\multirow{2}{*}{\textbf{Method}} &
  \multicolumn{3}{c|}{Playground} &
  \multicolumn{3}{c|}{Umbrella} &
  \multicolumn{3}{c|}{Skating} &
  \multicolumn{3}{c}{Mean} \\
 & PSNR $\uparrow$ & SSIM $\uparrow$ & LPIPS $\downarrow$ 
 & PSNR $\uparrow$ & SSIM $\uparrow$ & LPIPS $\downarrow$ 
 & PSNR $\uparrow$ & SSIM $\uparrow$ & LPIPS $\downarrow$ 
 & PSNR $\uparrow$ & SSIM $\uparrow$ & LPIPS $\downarrow$ \\
\midrule
DGM \cite{stearns2024dynamic} 
  & 16.68 & \textbf{0.60} & 0.177 
  & \textbf{24.99} & \textbf{0.70} & \textbf{0.083} 
  & 27.79 & 0.90 & 0.052 
  & \textbf{23.46} & \textbf{0.77} & \textbf{0.093} \\
Ours 
  & \textbf{16.87} & 0.55 & \textbf{0.168} 
  & 23.27 & 0.57 & 0.167 
  & \textbf{27.91} & \textbf{0.92} & \textbf{0.047} 
  & 23.37 & 0.75 & 0.102 \\
\bottomrule
\end{tabular}%
}
\caption{\textbf{Evaluation on NVIDIA dataset}. Our method is comparable with Dynamic Gaussian Marbles (DGM) \cite{stearns2024dynamic}.}
\label{tab:nvidia}
\end{table}

\subsection{Ablation studies}
We ablate various components of our method on the iPhone dataset in 3D tracking in Tab.~\ref{tab:ablations}.
We first validate our choices of motion representation, namely the $\mathbb{SE}(3)$ motion bases parameterization, with three ablations: 1) ``Per-Gaussian Transl.'': 
replacing our motion representation with naive per-Gaussian translational motion trajectories, 
2) ``Per-Gaussian $\SETHREE$'': 
replacing our motion representation with naive per-Gaussian $\SETHREE$ motion trajectories,
and 2) ``Transl. Bases'': keeping the motion bases representation but only using translational bases instead of $\mathbb{SE}(3)$.  
We find that $\mathbb{SE}(3)$ bases significantly improve 3D tracking over both translational bases and per-Gaussian translational motion. It also outperforms per-Gaussian $\mathbb{SE}(3)$ in overall accuracy and offers noticeably better visual quality, as the latter exhibits  significantly more popping and jitter artifacts. 

Next, we ablate our training strategies including initialization and supervision signal. We conduct an ablation of ``No $\mathbb{SE}(3)$ Init.'', where instead of performing our initial $\mathbb{SE}(3)$ fitting stage, we initialize the translational part of the motion bases with randomly selected noisy 3D tracks formed by directly lifting input 2D tracks using depth maps into 3D, and the rotation part as identity.
We find that skipping this initialization noticeably hurts performance. Lastly, we remove the 2D track supervision entirely (``No 2D Tracks'') and find it to lead to significant drop in performance, which verifies the importance of the 2D track supervision for 3D tracking.

\begin{table}[t]
    \centering
    \setlength{\tabcolsep}{0.4em} %
    \resizebox{\linewidth}{!}{%
    \begin{tabular}{l|cccc|ccc}
        \toprule
        \textbf{Methods} & $\mathbb{SE}(3)$ & Motion Basis & 2D tracks & Initialization  & EPE$\downarrow$ & $\deltathreedsmall\uparrow$ & $\deltathreedlarge\uparrow$ \\
        \midrule
        Ours (Full) & \checkmark & \checkmark & \checkmark & \checkmark & \textbf{0.082}  & 43.0 & \textbf{73.3} \\
        Transl. Bases &  &  \checkmark & \checkmark & \checkmark & 0.093  & 42.3 & 69.9 \\
        Per-Gaussian $\SETHREE$ & \checkmark &  & \checkmark & \checkmark & 0.083 & \textbf{43.6} & 70.2 \\
        Per-Gaussian Transl. &  &  & \checkmark & \checkmark & 0.087  & 41.2 & 69.2 \\
        No $\mathbb{SE}(3)$ Init. &  \checkmark & \checkmark & \checkmark &  & 0.111  & 39.3 & 65.7 \\
        No 2D Tracks & \checkmark & \checkmark &  &  & 0.141   & 30.4 & 57.8 \\
        \bottomrule
    \end{tabular}%
    }
    \caption{\textbf{Ablation Studies on iPhone dataset.}}
    \vspace{-2em}
    \label{tab:ablations}
\end{table}

\section{Discussion}
\label{sec:conclusion}

\noindent \textbf{Limitations.} 
Like most prior monocular view synthesis methods, it still requires per-scene test-time optimization, hindering streamable applications. Recent feed-forward methods for joint reconstruction and tracking~~\cite{feng2025st4rtrack,xiao2025spatialtrackerv2,xu20254dgt} offer a promising alternative, showing encouraging progress toward overcoming this limitation.
Our method relies on off-the-shelf predictions for camera poses, geometry, and motion, which may degrade in textureless regions or under large motions. However, it also naturally benefits from advances in these components.
Finally, it also requires user input to mask moving objects. Recent advances in moving object segmentation~\cite{goli2024romorobustmotionsegmentation, huang2025segmentmotionvideos} offer promising directions beyond our current manual solution and could be integrated into our pipeline.

\vspace{0.1em}
\noindent \textbf{Conclusion.}
We introduce a method for joint long-range 3D tracking and novel view synthesis from monocular video, representing dynamic elements via global 3D Gaussians with time-varying translations and rotations. Motion trajectories are regularized using low-dimensional rigid motion bases. By consolidating noisy observations into globally consistent scene estimates, our approach outperforms state-of-the-art methods in 2D/3D tracking and novel view synthesis across synthetic and real benchmarks.

\vspace{1em}
\noindent \textbf{Acknowledgement.} 
We thank Ruilong Li, Noah Snavely, Brent Yi and Aleksander Holynski for helpful discussion. 
We are in memory of our beloved cat Sriracha, who will always be missed and loved.
This project is supported in part by DARPA No. HR001123C0021. and  IARPA DOI/IBC No. 140D0423C0035.  The
views and conclusions contained herein are those of the authors and do not
represent the official policies or endorsements of these institutions.
{
    \small
    \bibliographystyle{ieeenat_fullname}
    \bibliography{main}

\begin{thebibliography}{130}
\providecommand{\natexlab}[1]{#1}
\providecommand{\url}[1]{\texttt{#1}}
\expandafter\ifx\csname urlstyle\endcsname\relax
  \providecommand{\doi}[1]{doi: #1}\else
  \providecommand{\doi}{doi: \begingroup \urlstyle{rm}\Url}\fi

\bibitem[Bansal et~al.(2020)Bansal, Vo, Sheikh, Ramanan, and
  Narasimhan]{bansal20204d}
Aayush Bansal, Minh Vo, Yaser Sheikh, Deva Ramanan, and Srinivasa Narasimhan.
\newblock 4d visualization of dynamic events from unconstrained multi-view
  videos.
\newblock In \emph{Proceedings of the IEEE/CVF Conference on Computer Vision
  and Pattern Recognition}, 2020.

\bibitem[Bay et~al.(2006)Bay, Tuytelaars, and Gool]{Bay2006SURFSU}
Herbert Bay, Tinne Tuytelaars, and Luc~Van Gool.
\newblock Surf: Speeded up robust features.
\newblock In \emph{European Conference on Computer Vision}, 2006.

\bibitem[Birchfield and Pundlik(2008)]{birchfield2008joint}
Stanley~T Birchfield and Shrinivas~J Pundlik.
\newblock Joint tracking of features and edges.
\newblock In \emph{2008 IEEE Conference on Computer Vision and Pattern
  Recognition}. IEEE, 2008.

\bibitem[Black and Anandan(1993)]{black1993framework}
Michael~J Black and Padmanabhan Anandan.
\newblock A framework for the robust estimation of optical flow.
\newblock In \emph{1993 (4th) International Conference on Computer Vision},
  1993.

\bibitem[Bozic et~al.(2019)Bozic, Zollh{\"o}fer, Theobalt, and
  Nie{\ss}ner]{Bozic2019DeepDeformLN}
Aljaz Bozic, Michael Zollh{\"o}fer, Christian Theobalt, and Matthias
  Nie{\ss}ner.
\newblock Deepdeform: Learning non-rigid rgb-d reconstruction with
  semi-supervised data.
\newblock \emph{2020 IEEE/CVF Conference on Computer Vision and Pattern
  Recognition (CVPR)}, 2019.

\bibitem[Bregler et~al.(2000)Bregler, Hertzmann, and Biermann]{854941}
C. Bregler, A. Hertzmann, and H. Biermann.
\newblock Recovering non-rigid 3d shape from image streams.
\newblock In \emph{Proceedings IEEE Conference on Computer Vision and Pattern
  Recognition. CVPR 2000 (Cat. No.PR00662)}, 2000.

\bibitem[Brox et~al.(2004)Brox, Bruhn, Papenberg, and Weickert]{Brox2004HighAO}
Thomas Brox, Andr{\'e}s Bruhn, Nils Papenberg, and Joachim Weickert.
\newblock High accuracy optical flow estimation based on a theory for warping.
\newblock In \emph{European Conference on Computer Vision}, 2004.

\bibitem[Brox et~al.(2009)Brox, Bregler, and Malik]{Brox2009LargeDO}
Thomas Brox, Christoph Bregler, and Jitendra Malik.
\newblock Large displacement optical flow.
\newblock \emph{2009 IEEE Conference on Computer Vision and Pattern
  Recognition}, 2009.

\bibitem[Broxton et~al.(2020)Broxton, Flynn, Overbeck, Erickson, Hedman,
  Duvall, Dourgarian, Busch, Whalen, and Debevec]{broxton2020immersive}
Michael Broxton, John Flynn, Ryan Overbeck, Daniel Erickson, Peter Hedman,
  Matthew Duvall, Jason Dourgarian, Jay Busch, Matt Whalen, and Paul Debevec.
\newblock Immersive light field video with a layered mesh representation.
\newblock \emph{ACM Transactions on Graphics (TOG)}, 39\penalty0 (4), 2020.

\bibitem[Cao and Johnson(2023)]{cao2023hexplane}
Ang Cao and Justin Johnson.
\newblock Hexplane: A fast representation for dynamic scenes.
\newblock In \emph{Proceedings of the IEEE/CVF Conference on Computer Vision
  and Pattern Recognition}, 2023.

\bibitem[Chen et~al.(2021)Chen, Zheng, Black, Hilliges, and
  Geiger]{chen2021snarf}
Xu Chen, Yufeng Zheng, Michael~J Black, Otmar Hilliges, and Andreas Geiger.
\newblock Snarf: Differentiable forward skinning for animating non-rigid neural
  implicit shapes.
\newblock In \emph{Proceedings of the IEEE/CVF International Conference on
  Computer Vision}, 2021.

\bibitem[Chu et~al.(2024)Chu, Ke, and Fragkiadaki]{dreamscene4d}
Wen-Hsuan Chu, Lei Ke, and Katerina Fragkiadaki.
\newblock Dreamscene4d: Dynamic multi-object scene generation from monocular
  videos.
\newblock \emph{NeurIPS}, 2024.

\bibitem[Dai et~al.(2014)Dai, Li, and He]{10.1007/s11263-013-0684-2}
Yuchao Dai, Hongdong Li, and Mingyi He.
\newblock A simple prior-free method for non-rigid structure-from-motion
  factorization.
\newblock \emph{Int. J. Comput. Vision}, 107\penalty0 (2), 2014.

\bibitem[DeTone et~al.(2017)DeTone, Malisiewicz, and
  Rabinovich]{DeTone2017SuperPointSI}
Daniel DeTone, Tomasz Malisiewicz, and Andrew Rabinovich.
\newblock Superpoint: Self-supervised interest point detection and description.
\newblock \emph{2018 IEEE/CVF Conference on Computer Vision and Pattern
  Recognition Workshops (CVPRW)}, 2017.

\bibitem[Doersch et~al.(2022)Doersch, Gupta, Markeeva, Recasens, Smaira, Aytar,
  Carreira, Zisserman, and Yang]{doersch2022tap}
Carl Doersch, Ankush Gupta, Larisa Markeeva, Adri{\`a} Recasens, Lucas Smaira,
  Yusuf Aytar, Jo{\~a}o Carreira, Andrew Zisserman, and Yi Yang.
\newblock Tap-vid: A benchmark for tracking any point in a video.
\newblock \emph{Advances in Neural Information Processing Systems}, 2022.

\bibitem[Doersch et~al.(2023)Doersch, Yang, Vecerik, Gokay, Gupta, Aytar,
  Carreira, and Zisserman]{doersch2023tapir}
Carl Doersch, Yi Yang, Mel Vecerik, Dilara Gokay, Ankush Gupta, Yusuf Aytar,
  Joao Carreira, and Andrew Zisserman.
\newblock Tapir: Tracking any point with per-frame initialization and temporal
  refinement.
\newblock In \emph{Proceedings of the IEEE/CVF International Conference on
  Computer Vision}, 2023.

\bibitem[Doersch et~al.(2024)Doersch, Luc, Yang, Gokay, Koppula, Gupta,
  Heyward, Rocco, Goroshin, Carreira, et~al.]{doersch2024bootstap}
Carl Doersch, Pauline Luc, Yi Yang, Dilara Gokay, Skanda Koppula, Ankush Gupta,
  Joseph Heyward, Ignacio Rocco, Ross Goroshin, Joao Carreira, et~al.
\newblock Bootstap: Bootstrapped training for tracking-any-point.
\newblock In \emph{Proceedings of the Asian Conference on Computer Vision},
  2024.

\bibitem[Dosovitskiy et~al.(2015)Dosovitskiy, Fischer, Ilg, H{\"a}usser,
  Hazirbas, Golkov, van~der Smagt, Cremers, and Brox]{Dosovitskiy2015FlowNetLO}
Alexey Dosovitskiy, Philipp Fischer, Eddy Ilg, Philip H{\"a}usser, Caner
  Hazirbas, Vladimir Golkov, Patrick van~der Smagt, Daniel Cremers, and Thomas
  Brox.
\newblock Flownet: Learning optical flow with convolutional networks.
\newblock \emph{2015 IEEE International Conference on Computer Vision (ICCV)},
  2015.

\bibitem[Dou et~al.(2016)Dou, Khamis, Degtyarev, Davidson, Fanello, Kowdle,
  Orts, Rhemann, Kim, Taylor, Kohli, Tankovich, and Izadi]{Dou2016Fusion4D}
Mingsong Dou, S. Khamis, Yu.G. Degtyarev, Philip~L. Davidson, S. Fanello,
  Adarsh Kowdle, Sergio Orts, Christoph Rhemann, David Kim, Jonathan Taylor,
  Pushmeet Kohli, Vladimir Tankovich, and Shahram Izadi.
\newblock Fusion4d.
\newblock \emph{ACM Transactions on Graphics (TOG)}, 35, 2016.

\bibitem[Du et~al.(2021)Du, Zhang, Yu, Tenenbaum, and Wu]{du2021neural}
Yilun Du, Yinan Zhang, Hong-Xing Yu, Joshua~B Tenenbaum, and Jiajun Wu.
\newblock Neural radiance flow for 4d view synthesis and video processing.
\newblock In \emph{2021 IEEE/CVF International Conference on Computer Vision
  (ICCV)}. IEEE Computer Society, 2021.

\bibitem[Duan et~al.(2024)Duan, Wei, Dai, He, Chen, and Chen]{duan20244d}
Yuanxing Duan, Fangyin Wei, Qiyu Dai, Yuhang He, Wenzheng Chen, and Baoquan
  Chen.
\newblock 4d-rotor gaussian splatting: towards efficient novel view synthesis
  for dynamic scenes.
\newblock In \emph{ACM SIGGRAPH 2024 Conference Papers}, 2024.

\bibitem[Feng et~al.(2025)Feng, Zhang, Wang, Ye, Yu, Black, Darrell, and
  Kanazawa]{feng2025st4rtrack}
Haiwen Feng, Junyi Zhang, Qianqian Wang, Yufei Ye, Pengcheng Yu, Michael~J
  Black, Trevor Darrell, and Angjoo Kanazawa.
\newblock St4rtrack: Simultaneous 4d reconstruction and tracking in the world.
\newblock \emph{arXiv preprint arXiv:2504.13152}, 2025.

\bibitem[Fridovich-Keil et~al.(2023)Fridovich-Keil, Meanti, Warburg, Recht, and
  Kanazawa]{fridovich2023k}
Sara Fridovich-Keil, Giacomo Meanti, Frederik~Rahb{\ae}k Warburg, Benjamin
  Recht, and Angjoo Kanazawa.
\newblock K-planes: Explicit radiance fields in space, time, and appearance.
\newblock In \emph{Proceedings of the IEEE/CVF Conference on Computer Vision
  and Pattern Recognition}, 2023.

\bibitem[Gao et~al.(2021)Gao, Saraf, Kopf, and Huang]{Gao-ICCV-DynNeRF}
Chen Gao, Ayush Saraf, Johannes Kopf, and Jia-Bin Huang.
\newblock Dynamic view synthesis from dynamic monocular video.
\newblock In \emph{Proceedings of the IEEE International Conference on Computer
  Vision}, 2021.

\bibitem[Gao et~al.(2022)Gao, Li, Tulsiani, Russell, and
  Kanazawa]{gao2022dynamic}
Hang Gao, Ruilong Li, Shubham Tulsiani, Bryan Russell, and Angjoo Kanazawa.
\newblock Dynamic novel-view synthesis: A reality check.
\newblock In \emph{NeurIPS}, 2022.

\bibitem[Gatys et~al.(2017)Gatys, Ecker, Bethge, Hertzmann, and
  Shechtman]{gatys2017controlling}
Leon~A Gatys, Alexander~S Ecker, Matthias Bethge, Aaron Hertzmann, and Eli
  Shechtman.
\newblock Controlling perceptual factors in neural style transfer.
\newblock In \emph{Proceedings of the IEEE conference on computer vision and
  pattern recognition}, 2017.

\bibitem[Goli et~al.(2024)Goli, Sabour, Matthews, Brubaker, Lagun, Jacobson,
  Fleet, Saxena, and Tagliasacchi]{goli2024romorobustmotionsegmentation}
Lily Goli, Sara Sabour, Mark Matthews, Marcus Brubaker, Dmitry Lagun, Alec
  Jacobson, David~J Fleet, Saurabh Saxena, and Andrea Tagliasacchi.
\newblock Romo: Robust motion segmentation improves structure from motion.
\newblock \emph{arXiv preprint arXiv:2411.18650}, 2024.

\bibitem[Greff et~al.(2022)Greff, Belletti, Beyer, Doersch, Du, Duckworth,
  Fleet, Gnanapragasam, Golemo, Herrmann, et~al.]{greff2022kubric}
Klaus Greff, Francois Belletti, Lucas Beyer, Carl Doersch, Yilun Du, Daniel
  Duckworth, David~J Fleet, Dan Gnanapragasam, Florian Golemo, Charles
  Herrmann, et~al.
\newblock Kubric: A scalable dataset generator.
\newblock In \emph{Proceedings of the IEEE/CVF Conference on Computer Vision
  and Pattern Recognition}, 2022.

\bibitem[Gu et~al.(2019)Gu, Wang, Wu, Lee, and Wang]{Gu2019HPLFlowNetHP}
Xiuye Gu, Yijie Wang, Chongruo Wu, Yong~Jae Lee, and Panqu Wang.
\newblock Hplflownet: Hierarchical permutohedral lattice flownet for scene flow
  estimation on large-scale point clouds.
\newblock \emph{2019 IEEE/CVF Conference on Computer Vision and Pattern
  Recognition (CVPR)}, 2019.

\bibitem[Harley et~al.(2022)Harley, Fang, and Fragkiadaki]{harley2022particle}
Adam~W Harley, Zhaoyuan Fang, and Katerina Fragkiadaki.
\newblock Particle video revisited: Tracking through occlusions using point
  trajectories.
\newblock In \emph{European Conference on Computer Vision}, 2022.

\bibitem[He et~al.(2024)He, Li, Yin, Liang, Li, Zhou, Zhang, Liu, and
  Chen]{he2024lotusdiffusionbasedvisualfoundation}
Jing He, Haodong Li, Wei Yin, Yixun Liang, Leheng Li, Kaiqiang Zhou, Hongbo
  Zhang, Bingbing Liu, and Ying-Cong Chen.
\newblock Lotus: Diffusion-based visual foundation model for high-quality dense
  prediction, 2024.

\bibitem[Hempel et~al.(2022)Hempel, Abdelrahman, and Al-Hamadi]{hempel20226d}
Thorsten Hempel, Ahmed~A Abdelrahman, and Ayoub Al-Hamadi.
\newblock 6d rotation representation for unconstrained head pose estimation.
\newblock In \emph{2022 IEEE International Conference on Image Processing
  (ICIP)}, 2022.

\bibitem[Horn and Schunck(1981)]{Horn1981DeterminingOF}
Berthold K.~P. Horn and Brian~G. Schunck.
\newblock Determining optical flow.
\newblock In \emph{Other Conferences}, 1981.

\bibitem[Huang et~al.(2024)Huang, Yu, Chen, Geiger, and Gao]{Huang2DGS2024}
Binbin Huang, Zehao Yu, Anpei Chen, Andreas Geiger, and Shenghua Gao.
\newblock 2d gaussian splatting for geometrically accurate radiance fields.
\newblock In \emph{SIGGRAPH 2024 Conference Papers}. Association for Computing
  Machinery, 2024.

\bibitem[Huang et~al.(2025)Huang, Zheng, Xu, Keutzer, Zhang, Kanazawa, and
  Wang]{huang2025segmentmotionvideos}
Nan Huang, Wenzhao Zheng, Chenfeng Xu, Kurt Keutzer, Shanghang Zhang, Angjoo
  Kanazawa, and Qianqian Wang.
\newblock Segment any motion in videos.
\newblock In \emph{Proceedings of the Computer Vision and Pattern Recognition
  Conference (CVPR)}, 2025.

\bibitem[Huang et~al.(2022)Huang, Shi, Zhang, Wang, Cheung, Qin, Dai, and
  Li]{huang2022flowformer}
Zhaoyang Huang, Xiaoyu Shi, Chao Zhang, Qiang Wang, Ka~Chun Cheung, Hongwei
  Qin, Jifeng Dai, and Hongsheng Li.
\newblock Flowformer: A transformer architecture for optical flow.
\newblock In \emph{European Conference on Computer Vision}. Springer, 2022.

\bibitem[Huh et~al.(2020)Huh, Zhang, Zhu, Paris, and
  Hertzmann]{huh2020transforming}
Minyoung Huh, Richard Zhang, Jun-Yan Zhu, Sylvain Paris, and Aaron Hertzmann.
\newblock Transforming and projecting images into class-conditional generative
  networks.
\newblock In \emph{Computer Vision--ECCV 2020: 16th European Conference,
  Glasgow, UK, August 23--28, 2020, Proceedings, Part II 16}, 2020.

\bibitem[Hur and Roth(2020)]{Hur2020SelfSupervisedMS}
Junhwa Hur and Stefan Roth.
\newblock Self-supervised monocular scene flow estimation.
\newblock \emph{2020 IEEE/CVF Conference on Computer Vision and Pattern
  Recognition (CVPR)}, 2020.

\bibitem[I\c{s}{\i}k et~al.(2023)I\c{s}{\i}k, Rünz, Georgopoulos, Khakhulin,
  Starck, Agapito, and Nießner]{isik2023humanrf}
Mustafa I\c{s}{\i}k, Martin Rünz, Markos Georgopoulos, Taras Khakhulin,
  Jonathan Starck, Lourdes Agapito, and Matthias Nießner.
\newblock Humanrf: High-fidelity neural radiance fields for humans in motion.
\newblock \emph{ACM Transactions on Graphics (TOG)}, 2023.

\bibitem[Ilg et~al.(2017)Ilg, Mayer, Saikia, Keuper, Dosovitskiy, and
  Brox]{ilg2017flownet}
Eddy Ilg, Nikolaus Mayer, Tonmoy Saikia, Margret Keuper, Alexey Dosovitskiy,
  and Thomas Brox.
\newblock Flownet 2.0: Evolution of optical flow estimation with deep networks.
\newblock In \emph{Proceedings of the IEEE conference on computer vision and
  pattern recognition}, 2017.

\bibitem[Innmann et~al.(2016)Innmann, Zollh{\"o}fer, Nie{\ss}ner, Theobalt, and
  Stamminger]{Innmann2016VolumeDeformRV}
Matthias Innmann, Michael Zollh{\"o}fer, Matthias Nie{\ss}ner, Christian
  Theobalt, and Marc Stamminger.
\newblock Volumedeform: Real-time volumetric non-rigid reconstruction.
\newblock In \emph{European Conference on Computer Vision}, 2016.

\bibitem[Janai et~al.(2018)Janai, Guney, Ranjan, Black, and
  Geiger]{janai2018unsupervised}
Joel Janai, Fatma Guney, Anurag Ranjan, Michael Black, and Andreas Geiger.
\newblock Unsupervised learning of multi-frame optical flow with occlusions.
\newblock In \emph{Proceedings of the European conference on computer vision
  (ECCV)}, 2018.

\bibitem[Jiang et~al.(2021{\natexlab{a}})Jiang, Campbell, Lu, Li, and
  Hartley]{jiang2021learning}
Shihao Jiang, Dylan Campbell, Yao Lu, Hongdong Li, and Richard Hartley.
\newblock Learning to estimate hidden motions with global motion aggregation.
\newblock In \emph{Proceedings of the IEEE/CVF International Conference on
  Computer Vision}, 2021{\natexlab{a}}.

\bibitem[Jiang et~al.(2021{\natexlab{b}})Jiang, Lu, Li, and
  Hartley]{jiang2021learningf}
Shihao Jiang, Yao Lu, Hongdong Li, and Richard Hartley.
\newblock Learning optical flow from a few matches.
\newblock In \emph{Proceedings of the IEEE/CVF conference on computer vision
  and pattern recognition}, 2021{\natexlab{b}}.

\bibitem[Karaev et~al.(2024)Karaev, Rocco, Graham, Neverova, Vedaldi, and
  Rupprecht]{karaev2023cotracker}
Nikita Karaev, Ignacio Rocco, Benjamin Graham, Natalia Neverova, Andrea
  Vedaldi, and Christian Rupprecht.
\newblock Cotracker: It is better to track together.
\newblock In \emph{European conference on computer vision}, 2024.

\bibitem[Kerbl et~al.(2023{\natexlab{a}})Kerbl, Kopanas, Leimk{\"u}hler, and
  Drettakis]{kerbl20233d}
Bernhard Kerbl, Georgios Kopanas, Thomas Leimk{\"u}hler, and George Drettakis.
\newblock 3d gaussian splatting for real-time radiance field rendering.
\newblock \emph{ACM Transactions on Graphics}, 2023{\natexlab{a}}.

\bibitem[Kerbl et~al.(2023{\natexlab{b}})Kerbl, Kopanas, Leimk{\"u}hler, and
  Drettakis]{kerbl3Dgaussians}
Bernhard Kerbl, Georgios Kopanas, Thomas Leimk{\"u}hler, and George Drettakis.
\newblock 3d gaussian splatting for real-time radiance field rendering.
\newblock \emph{ACM Transactions on Graphics}, 2023{\natexlab{b}}.

\bibitem[Kingma and Ba(2014)]{kingma2014adam}
Diederik~P Kingma and Jimmy Ba.
\newblock Adam: A method for stochastic optimization.
\newblock \emph{arXiv preprint arXiv:1412.6980}, 2014.

\bibitem[Kingma et~al.(2020)Kingma, Ba, and Adam]{kingma2020method}
Diederik~P Kingma, J~Adam Ba, and J Adam.
\newblock A method for stochastic optimization. arxiv 2014.
\newblock \emph{arXiv preprint arXiv:1412.6980}, 2020.

\bibitem[Kirillov et~al.(2023)Kirillov, Mintun, Ravi, Mao, Rolland, Gustafson,
  Xiao, Whitehead, Berg, Lo, Doll{\'a}r, and Girshick]{Kirillov2023SegmentA}
Alexander Kirillov, Eric Mintun, Nikhila Ravi, Hanzi Mao, Chloe Rolland, Laura
  Gustafson, Tete Xiao, Spencer Whitehead, Alexander~C. Berg, Wan-Yen Lo, Piotr
  Doll{\'a}r, and Ross~B. Girshick.
\newblock Segment anything.
\newblock \emph{2023 IEEE/CVF International Conference on Computer Vision
  (ICCV)}, 2023.

\bibitem[Kopf et~al.(2021)Kopf, Rong, and Huang]{kopf2021robust}
Johannes Kopf, Xuejian Rong, and Jia-Bin Huang.
\newblock Robust consistent video depth estimation.
\newblock In \emph{Proceedings of the IEEE/CVF Conference on Computer Vision
  and Pattern Recognition}, 2021.

\bibitem[Koppula et~al.(2024)Koppula, Rocco, Yang, Heyward, Carreira,
  Zisserman, Brostow, and Doersch]{koppula2024tapvid}
Skanda Koppula, Ignacio Rocco, Yi Yang, Joe Heyward, Joao Carreira, Andrew
  Zisserman, Gabriel Brostow, and Carl Doersch.
\newblock Tapvid-3d: A benchmark for tracking any point in 3d.
\newblock \emph{Advances in Neural Information Processing Systems}, 2024.

\bibitem[Kratimenos et~al.(2024)Kratimenos, Lei, and
  Daniilidis]{kratimenos2024dynmf}
Agelos Kratimenos, Jiahui Lei, and Kostas Daniilidis.
\newblock Dynmf: Neural motion factorization for real-time dynamic view
  synthesis with 3d gaussian splatting.
\newblock \emph{ECCV}, 2024.

\bibitem[Kumar et~al.(2017)Kumar, Dai, and Li]{Kumar2017MonocularD3}
Suryansh Kumar, Yuchao Dai, and Hongdong Li.
\newblock Monocular dense 3d reconstruction of a complex dynamic scene from two
  perspective frames.
\newblock \emph{2017 IEEE International Conference on Computer Vision (ICCV)},
  2017.

\bibitem[Lei et~al.(2025)Lei, Weng, Harley, Guibas, and
  Daniilidis]{lei2024mosca}
Jiahui Lei, Yijia Weng, Adam~W Harley, Leonidas Guibas, and Kostas Daniilidis.
\newblock Mosca: Dynamic gaussian fusion from casual videos via 4d motion
  scaffolds.
\newblock In \emph{Proceedings of the Computer Vision and Pattern Recognition
  Conference}, 2025.

\bibitem[Li et~al.(2022{\natexlab{a}})Li, Tanke, Vo, Zollh{\"o}fer, Gall,
  Kanazawa, and Lassner]{li2022tava}
Ruilong Li, Julian Tanke, Minh Vo, Michael Zollh{\"o}fer, J{\"u}rgen Gall,
  Angjoo Kanazawa, and Christoph Lassner.
\newblock Tava: Template-free animatable volumetric actors.
\newblock In \emph{European Conference on Computer Vision}, 2022{\natexlab{a}}.

\bibitem[Li et~al.(2022{\natexlab{b}})Li, Slavcheva, Zollhoefer, Green,
  Lassner, Kim, Schmidt, Lovegrove, Goesele, Newcombe, et~al.]{li2022neural}
Tianye Li, Mira Slavcheva, Michael Zollhoefer, Simon Green, Christoph Lassner,
  Changil Kim, Tanner Schmidt, Steven Lovegrove, Michael Goesele, Richard
  Newcombe, et~al.
\newblock Neural 3d video synthesis from multi-view video.
\newblock In \emph{Proceedings of the IEEE/CVF Conference on Computer Vision
  and Pattern Recognition}, 2022{\natexlab{b}}.

\bibitem[Li et~al.(2019)Li, Dekel, Cole, Tucker, Snavely, Liu, and
  Freeman]{li2019learning}
Zhengqi Li, Tali Dekel, Forrester Cole, Richard Tucker, Noah Snavely, Ce Liu,
  and William~T Freeman.
\newblock Learning the depths of moving people by watching frozen people.
\newblock In \emph{Proceedings of the IEEE/CVF conference on computer vision
  and pattern recognition}, 2019.

\bibitem[Li et~al.(2021)Li, Niklaus, Snavely, and Wang]{li2021neural}
Zhengqi Li, Simon Niklaus, Noah Snavely, and Oliver Wang.
\newblock Neural scene flow fields for space-time view synthesis of dynamic
  scenes.
\newblock In \emph{Proceedings of the IEEE/CVF Conference on Computer Vision
  and Pattern Recognition}, 2021.

\bibitem[Li et~al.(2023)Li, Wang, Cole, Tucker, and Snavely]{li2023dynibar}
Zhengqi Li, Qianqian Wang, Forrester Cole, Richard Tucker, and Noah Snavely.
\newblock Dynibar: Neural dynamic image-based rendering.
\newblock In \emph{Proceedings of the IEEE/CVF Conference on Computer Vision
  and Pattern Recognition}, 2023.

\bibitem[Li et~al.(2024)Li, Chen, Li, and Xu]{li2023spacetime}
Zhan Li, Zhang Chen, Zhong Li, and Yi Xu.
\newblock Spacetime gaussian feature splatting for real-time dynamic view
  synthesis.
\newblock In \emph{Proceedings of the IEEE/CVF Conference on Computer Vision
  and Pattern Recognition}, 2024.

\bibitem[Li et~al.(2025)Li, Tucker, Cole, Wang, Jin, Ye, Kanazawa, Holynski,
  and Snavely]{li2024_MegaSaM}
Zhengqi Li, Richard Tucker, Forrester Cole, Qianqian Wang, Linyi Jin, Vickie
  Ye, Angjoo Kanazawa, Aleksander Holynski, and Noah Snavely.
\newblock Megasam: Accurate, fast and robust structure and motion from casual
  dynamic videos.
\newblock In \emph{Proceedings of the Computer Vision and Pattern Recognition
  Conference}, 2025.

\bibitem[Liu et~al.(2011)Liu, Yuen, and Torralba]{Liu2011SIFTFD}
Ce Liu, Jenny Yuen, and Antonio Torralba.
\newblock Sift flow: Dense correspondence across scenes and its applications.
\newblock \emph{IEEE Transactions on Pattern Analysis and Machine
  Intelligence}, 2011.

\bibitem[Liu et~al.(2025)Liu, Liu, Wang, Lv, Wang, Wang, and Hou]{liu2024modgs}
Qingming Liu, Yuan Liu, Jiepeng Wang, Xianqiang Lv, Peng Wang, Wenping Wang,
  and Junhui Hou.
\newblock Modgs: Dynamic gaussian splatting from causually-captured monocular
  videos.
\newblock \emph{ICLR}, 2025.

\bibitem[Liu et~al.(2019)Liu, Qi, and Guibas]{Liu2018LearningSF}
Xingyu Liu, Charles~R Qi, and Leonidas~J Guibas.
\newblock Flownet3d: Learning scene flow in 3d point clouds.
\newblock In \emph{Proceedings of the IEEE/CVF conference on computer vision
  and pattern recognition}, 2019.

\bibitem[Lowe(2004)]{Lowe2004DistinctiveIF}
David~G. Lowe.
\newblock Distinctive image features from scale-invariant keypoints.
\newblock \emph{International Journal of Computer Vision}, 2004.

\bibitem[Lu et~al.(2025)Lu, Huang, Li, Dou, Lin, Cui, Dong, Yeung, Wang, and
  Liu]{lu2025align3r}
Jiahao Lu, Tianyu Huang, Peng Li, Zhiyang Dou, Cheng Lin, Zhiming Cui, Zhen
  Dong, Sai-Kit Yeung, Wenping Wang, and Yuan Liu.
\newblock Align3r: Aligned monocular depth estimation for dynamic videos.
\newblock In \emph{Proceedings of the Computer Vision and Pattern Recognition
  Conference}, 2025.

\bibitem[Lucas and Kanade(1981)]{Lucas1981AnII}
Bruce~D. Lucas and Takeo Kanade.
\newblock An iterative image registration technique with an application to
  stereo vision.
\newblock In \emph{International Joint Conference on Artificial Intelligence},
  1981.

\bibitem[Luiten et~al.(2024)Luiten, Kopanas, Leibe, and
  Ramanan]{Luiten2023Dynamic3G}
Jonathon Luiten, Georgios Kopanas, Bastian Leibe, and Deva Ramanan.
\newblock Dynamic 3d gaussians: Tracking by persistent dynamic view synthesis.
\newblock In \emph{2024 International Conference on 3D Vision (3DV)}, 2024.

\bibitem[Luo et~al.(2020)Luo, Huang, Szeliski, Matzen, and
  Kopf]{luo2020consistent}
Xuan Luo, Jia-Bin Huang, Richard Szeliski, Kevin Matzen, and Johannes Kopf.
\newblock Consistent video depth estimation.
\newblock \emph{ACM Transactions on Graphics (ToG)}, 39\penalty0 (4), 2020.

\bibitem[Mildenhall et~al.(2020)Mildenhall, Srinivasan, Tancik, Barron,
  Ramamoorthi, and Ng]{mildenhall2020nerf}
Ben Mildenhall, Pratul~P. Srinivasan, Matthew Tancik, Jonathan~T. Barron, Ravi
  Ramamoorthi, and Ren Ng.
\newblock Nerf: Representing scenes as neural radiance fields for view
  synthesis.
\newblock In \emph{ECCV}, 2020.

\bibitem[Neoral et~al.(2024)Neoral, \v{S}er\'ych, and Matas]{Neoral_2024_WACV}
Michal Neoral, Jon\'a\v{s} \v{S}er\'ych, and Ji\v{r}{\'\i} Matas.
\newblock Mft: Long-term tracking of every pixel.
\newblock In \emph{Proceedings of the IEEE/CVF Winter Conference on
  Applications of Computer Vision (WACV)}, 2024.

\bibitem[Newcombe et~al.(2015)Newcombe, Fox, and
  Seitz]{Newcombe2015DynamicFusionRA}
Richard~A. Newcombe, Dieter Fox, and Steven~M. Seitz.
\newblock Dynamicfusion: Reconstruction and tracking of non-rigid scenes in
  real-time.
\newblock \emph{2015 IEEE Conference on Computer Vision and Pattern Recognition
  (CVPR)}, 2015.

\bibitem[Ngo et~al.(2025)Ngo, Zhuang, Gan, Kalogerakis, Tulyakov, Lee, and
  Wang]{ngo2025deltadenseefficientlongrange}
Tuan~Duc Ngo, Peiye Zhuang, Chuang Gan, Evangelos Kalogerakis, Sergey Tulyakov,
  Hsin-Ying Lee, and Chaoyang Wang.
\newblock Delta: Dense efficient long-range 3d tracking for any video.
\newblock In \emph{ICLR}, 2025.

\bibitem[Novotny et~al.(2019)Novotny, Ravi, Graham, Neverova, and
  Vedaldi]{novotny2019c3dpocanonical3dpose}
David Novotny, Nikhila Ravi, Benjamin Graham, Natalia Neverova, and Andrea
  Vedaldi.
\newblock C3dpo: Canonical 3d pose networks for non-rigid structure from
  motion.
\newblock In \emph{ICCV}, 2019.

\bibitem[Park et~al.(2021{\natexlab{a}})Park, Sinha, Barron, Bouaziz, Goldman,
  Seitz, and Martin-Brualla]{park2021nerfies}
Keunhong Park, Utkarsh Sinha, Jonathan~T Barron, Sofien Bouaziz, Dan~B Goldman,
  Steven~M Seitz, and Ricardo Martin-Brualla.
\newblock Nerfies: Deformable neural radiance fields.
\newblock In \emph{Proceedings of the IEEE/CVF International Conference on
  Computer Vision}, 2021{\natexlab{a}}.

\bibitem[Park et~al.(2021{\natexlab{b}})Park, Sinha, Hedman, Barron, Bouaziz,
  Goldman, Martin-Brualla, and Seitz]{park2021hypernerf}
Keunhong Park, Utkarsh Sinha, Peter Hedman, Jonathan~T Barron, Sofien Bouaziz,
  Dan~B Goldman, Ricardo Martin-Brualla, and Steven~M Seitz.
\newblock Hypernerf: A higher-dimensional representation for topologically
  varying neural radiance fields.
\newblock \emph{SIGGRAPH Asia}, 2021{\natexlab{b}}.

\bibitem[Piccinelli et~al.(2024)Piccinelli, Yang, Sakaridis, Segu, Li,
  Van~Gool, and Yu]{piccinelli2024unidepth}
Luigi Piccinelli, Yung-Hsu Yang, Christos Sakaridis, Mattia Segu, Siyuan Li,
  Luc Van~Gool, and Fisher Yu.
\newblock {U}ni{D}epth: Universal monocular metric depth estimation.
\newblock In \emph{Proceedings of the IEEE/CVF Conference on Computer Vision
  and Pattern Recognition (CVPR)}, 2024.

\bibitem[Pumarola et~al.(2021)Pumarola, Corona, Pons-Moll, and
  Moreno-Noguer]{pumarola2021d}
Albert Pumarola, Enric Corona, Gerard Pons-Moll, and Francesc Moreno-Noguer.
\newblock D-nerf: Neural radiance fields for dynamic scenes.
\newblock In \emph{Proceedings of the IEEE/CVF Conference on Computer Vision
  and Pattern Recognition}, 2021.

\bibitem[Puy et~al.(2020)Puy, Boulch, and Marlet]{Puy2020FLOTSF}
Gilles Puy, Alexandre Boulch, and Renaud Marlet.
\newblock Flot: Scene flow on point clouds guided by optimal transport.
\newblock In \emph{European Conference on Computer Vision}, 2020.

\bibitem[Ranftl et~al.(2016)Ranftl, Vineet, Chen, and Koltun]{Ranftl_2016_CVPR}
Rene Ranftl, Vibhav Vineet, Qifeng Chen, and Vladlen Koltun.
\newblock Dense monocular depth estimation in complex dynamic scenes.
\newblock In \emph{Proceedings of the IEEE Conference on Computer Vision and
  Pattern Recognition (CVPR)}, 2016.

\bibitem[Ren et~al.(2019)Ren, Gallo, Sun, Yang, Sudderth, and
  Kautz]{ren2019fusion}
Zhile Ren, Orazio Gallo, Deqing Sun, Ming-Hsuan Yang, Erik~B Sudderth, and Jan
  Kautz.
\newblock A fusion approach for multi-frame optical flow estimation.
\newblock In \emph{2019 IEEE Winter Conference on Applications of Computer
  Vision (WACV)}, 2019.

\bibitem[Rubinstein and Liu(2012)]{Rubinstein2012TowardsLL}
Michael Rubinstein and Ce Liu.
\newblock Towards longer long-range motion trajectories.
\newblock In \emph{British Machine Vision Conference}, 2012.

\bibitem[Rublee et~al.(2011)Rublee, Rabaud, Konolige, and
  Bradski]{Rublee2011ORBAE}
Ethan Rublee, Vincent Rabaud, Kurt Konolige, and Gary~R. Bradski.
\newblock Orb: An efficient alternative to sift or surf.
\newblock \emph{2011 International Conference on Computer Vision}, 2011.

\bibitem[Russell et~al.(2014)Russell, Yu, and Agapito]{russell2014video}
Chris Russell, Rui Yu, and Lourdes Agapito.
\newblock Video pop-up: Monocular 3d reconstruction of dynamic scenes.
\newblock In \emph{European conference on computer vision}. Springer, 2014.

\bibitem[Sand(2006)]{sandlong}
P Sand.
\newblock \emph{Long-range video motion estimation using point trajectories}.
\newblock PhD thesis, Ph. D. dissertation, Cambridge, MA, USA, 2006,
  adviser-Teller, Seth, 2006.

\bibitem[Sand and Teller(2008)]{sand2008particle}
Peter Sand and Seth Teller.
\newblock Particle video: Long-range motion estimation using point
  trajectories.
\newblock \emph{International journal of computer vision}, 2008.

\bibitem[Sch\"{o}nberger and Frahm(2016)]{schoenberger2016sfm}
Johannes~Lutz Sch\"{o}nberger and Jan-Michael Frahm.
\newblock Structure-from-motion revisited.
\newblock In \emph{Conference on Computer Vision and Pattern Recognition
  (CVPR)}, 2016.

\bibitem[Schonberger and Frahm(2016)]{schonberger2016structure}
Johannes~L Schonberger and Jan-Michael Frahm.
\newblock Structure-from-motion revisited.
\newblock In \emph{Proceedings of the IEEE conference on computer vision and
  pattern recognition}, 2016.

\bibitem[Shi et~al.(2023)Shi, Huang, Bian, Li, Zhang, Cheung, See, Qin, Dai,
  and Li]{shi2023videoflow}
Xiaoyu Shi, Zhaoyang Huang, Weikang Bian, Dasong Li, Manyuan Zhang, Ka~Chun
  Cheung, Simon See, Hongwei Qin, Jifeng Dai, and Hongsheng Li.
\newblock Videoflow: Exploiting temporal cues for multi-frame optical flow
  estimation.
\newblock In \emph{ICCV}, 2023.

\bibitem[Sivic et~al.(2004)Sivic, Schaffalitzky, and
  Zisserman]{Sivic2004ObjectLG}
Josef Sivic, Frederik Schaffalitzky, and Andrew Zisserman.
\newblock Object level grouping for video shots.
\newblock \emph{International Journal of Computer Vision}, 2004.

\bibitem[Song et~al.(2023)Song, Chen, Li, Chen, Chen, Yuan, Xu, and
  Geiger]{song2023nerfplayer}
Liangchen Song, Anpei Chen, Zhong Li, Zhang Chen, Lele Chen, Junsong Yuan, Yi
  Xu, and Andreas Geiger.
\newblock Nerfplayer: A streamable dynamic scene representation with decomposed
  neural radiance fields.
\newblock \emph{IEEE Transactions on Visualization and Computer Graphics},
  29\penalty0 (5), 2023.

\bibitem[Song et~al.(2024)Song, Lei, Wang, Liu, and Daniilidis]{song2024track}
Yunzhou Song, Jiahui Lei, Ziyun Wang, Lingjie Liu, and Kostas Daniilidis.
\newblock Track everything everywhere fast and robustly.
\newblock In \emph{European Conference on Computer Vision}, 2024.

\bibitem[Stearns et~al.(2024)Stearns, Harley, Uy, Dubost, Tombari, Wetzstein,
  and Guibas]{stearns2024dynamic}
Colton Stearns, Adam Harley, Mikaela Uy, Florian Dubost, Federico Tombari,
  Gordon Wetzstein, and Leonidas Guibas.
\newblock Dynamic gaussian marbles for novel view synthesis of casual monocular
  videos.
\newblock \emph{Siggraph Asia}, 2024.

\bibitem[Stich et~al.(2008)Stich, Linz, Albuquerque, and Magnor]{stich2008view}
Timo Stich, Christian Linz, Georgia Albuquerque, and Marcus Magnor.
\newblock View and time interpolation in image space.
\newblock In \emph{Computer Graphics Forum}. Wiley Online Library, 2008.

\bibitem[Sun et~al.(2017)Sun, Yang, Liu, and Kautz]{Sun2017PWCNetCF}
Deqing Sun, Xiaodong Yang, Ming-Yu Liu, and Jan Kautz.
\newblock Pwc-net: Cnns for optical flow using pyramid, warping, and cost
  volume.
\newblock \emph{2018 IEEE/CVF Conference on Computer Vision and Pattern
  Recognition}, 2017.

\bibitem[Teed and Deng(2020)]{teed2020raft}
Zachary Teed and Jia Deng.
\newblock Raft: Recurrent all-pairs field transforms for optical flow.
\newblock In \emph{Computer Vision--ECCV 2020: 16th European Conference,
  Glasgow, UK, August 23--28, 2020, Proceedings, Part II 16}. Springer, 2020.

\bibitem[Teed and Deng(2021{\natexlab{a}})]{teed2021droid}
Zachary Teed and Jia Deng.
\newblock Droid-slam: Deep visual slam for monocular, stereo, and rgb-d
  cameras.
\newblock \emph{Advances in neural information processing systems},
  2021{\natexlab{a}}.

\bibitem[Teed and Deng(2021{\natexlab{b}})]{teed2021raft3d}
Zachary Teed and Jia Deng.
\newblock Raft-3d: Scene flow using rigid-motion embeddings.
\newblock In \emph{Proceedings of the IEEE/CVF Conference on Computer Vision
  and Pattern Recognition (CVPR)}, 2021{\natexlab{b}}.

\bibitem[Wang et~al.(2021)Wang, Eckart, Lucey, and Gallo]{wang2021neural}
Chaoyang Wang, Ben Eckart, Simon Lucey, and Orazio Gallo.
\newblock Neural trajectory fields for dynamic novel view synthesis.
\newblock \emph{arXiv preprint arXiv:2105.05994}, 2021.

\bibitem[Wang et~al.(2022{\natexlab{a}})Wang, Li, Pontes, and
  Lucey]{Wang2022NeuralPF}
Chaoyang Wang, Xueqian Li, Jhony~Kaesemodel Pontes, and Simon Lucey.
\newblock Neural prior for trajectory estimation.
\newblock \emph{2022 IEEE/CVF Conference on Computer Vision and Pattern
  Recognition (CVPR)}, 2022{\natexlab{a}}.

\bibitem[Wang and Schmid(2013)]{Wang2013ActionRW}
Heng Wang and Cordelia Schmid.
\newblock Action recognition with improved trajectories.
\newblock \emph{2013 IEEE International Conference on Computer Vision}, 2013.

\bibitem[Wang et~al.(2022{\natexlab{b}})Wang, Zhang, Liu, Zhao, Zhang, Zhang,
  Wu, Yu, and Xu]{wang2022fourier}
Liao Wang, Jiakai Zhang, Xinhang Liu, Fuqiang Zhao, Yanshun Zhang, Yingliang
  Zhang, Minye Wu, Jingyi Yu, and Lan Xu.
\newblock Fourier plenoctrees for dynamic radiance field rendering in
  real-time.
\newblock In \emph{Proceedings of the IEEE/CVF Conference on Computer Vision
  and Pattern Recognition}, 2022{\natexlab{b}}.

\bibitem[Wang et~al.(2023)Wang, Chang, Cai, Li, Hariharan, Holynski, and
  Snavely]{wang2023omnimotion}
Qianqian Wang, Yen-Yu Chang, Ruojin Cai, Zhengqi Li, Bharath Hariharan,
  Aleksander Holynski, and Noah Snavely.
\newblock Tracking everything everywhere all at once.
\newblock In \emph{International Conference on Computer Vision}, 2023.

\bibitem[Wang et~al.(2025{\natexlab{a}})Wang, Zhang, Holynski, Efros, and
  Kanazawa]{wang2025continuous}
Qianqian Wang, Yifei Zhang, Aleksander Holynski, Alexei~A Efros, and Angjoo
  Kanazawa.
\newblock Continuous 3d perception model with persistent state.
\newblock In \emph{Proceedings of the Computer Vision and Pattern Recognition
  Conference}, 2025{\natexlab{a}}.

\bibitem[Wang et~al.(2024)Wang, Leroy, Cabon, Chidlovskii, and
  Revaud]{dust3r_cvpr24}
Shuzhe Wang, Vincent Leroy, Yohann Cabon, Boris Chidlovskii, and Jerome Revaud.
\newblock Dust3r: Geometric 3d vision made easy.
\newblock In \emph{CVPR}, 2024.

\bibitem[Wang et~al.(2025{\natexlab{b}})Wang, Yang, Shen, Jiang, and
  Wang]{wang2024gflow}
Shizun Wang, Xingyi Yang, Qiuhong Shen, Zhenxiang Jiang, and Xinchao Wang.
\newblock Gflow: Recovering 4d world from monocular video.
\newblock In \emph{Proceedings of the AAAI Conference on Artificial
  Intelligence}, 2025{\natexlab{b}}.

\bibitem[Wang et~al.(2004)Wang, Bovik, Sheikh, and Simoncelli]{wang2004image}
Zhou Wang, Alan~C Bovik, Hamid~R Sheikh, and Eero~P Simoncelli.
\newblock Image quality assessment: from error visibility to structural
  similarity.
\newblock \emph{IEEE transactions on image processing}, 2004.

\bibitem[Wang et~al.(2019)Wang, Li, Howard-Jenkins, Prisacariu, and
  Chen]{Wang2019FlowNet3DGL}
Zirui Wang, Shuda Li, Henry Howard-Jenkins, Victor~Adrian Prisacariu, and Min
  Chen.
\newblock Flownet3d++: Geometric losses for deep scene flow estimation.
\newblock \emph{2020 IEEE Winter Conference on Applications of Computer Vision
  (WACV)}, 2019.

\bibitem[Weng et~al.(2022)Weng, Curless, Srinivasan, Barron, and
  Kemelmacher-Shlizerman]{weng2022humannerf}
Chung-Yi Weng, Brian Curless, Pratul~P Srinivasan, Jonathan~T Barron, and Ira
  Kemelmacher-Shlizerman.
\newblock Humannerf: Free-viewpoint rendering of moving people from monocular
  video.
\newblock In \emph{Proceedings of the IEEE/CVF conference on computer vision
  and pattern Recognition}, 2022.

\bibitem[Wu et~al.(2023)Wu, Yi, Fang, Xie, Zhang, Wei, Liu, Tian, and
  Wang]{wu20234d}
Guanjun Wu, Taoran Yi, Jiemin Fang, Lingxi Xie, Xiaopeng Zhang, Wei Wei, Wenyu
  Liu, Qi Tian, and Xinggang Wang.
\newblock 4d gaussian splatting for real-time dynamic scene rendering.
\newblock In \emph{CVPR}, 2023.

\bibitem[Xian et~al.(2021)Xian, Huang, Kopf, and Kim]{xian2021space}
Wenqi Xian, Jia-Bin Huang, Johannes Kopf, and Changil Kim.
\newblock Space-time neural irradiance fields for free-viewpoint video.
\newblock In \emph{Proceedings of the IEEE/CVF Conference on Computer Vision
  and Pattern Recognition}, 2021.

\bibitem[Xiao et~al.(2024)Xiao, Wang, Zhang, Xue, Peng, Shen, and
  Zhou]{xiao2024spatialtrackertracking2dpixels}
Yuxi Xiao, Qianqian Wang, Shangzhan Zhang, Nan Xue, Sida Peng, Yujun Shen, and
  Xiaowei Zhou.
\newblock Spatialtracker: Tracking any 2d pixels in 3d space.
\newblock In \emph{Proceedings of the IEEE/CVF Conference on Computer Vision
  and Pattern Recognition}, 2024.

\bibitem[Xiao et~al.(2025)Xiao, Wang, Xue, Karaev, Makarov, Kang, Zhu, Bao,
  Shen, and Zhou]{xiao2025spatialtrackerv2}
Yuxi Xiao, Jianyuan Wang, Nan Xue, Nikita Karaev, Yuri Makarov, Bingyi Kang,
  Xing Zhu, Hujun Bao, Yujun Shen, and Xiaowei Zhou.
\newblock Spatialtrackerv2: 3d point tracking made easy.
\newblock In \emph{Proceedings of the IEEE/CVF International Conference on
  Computer Vision}, 2025.

\bibitem[Xu et~al.(2022)Xu, Zhang, Cai, Rezatofighi, and Tao]{xu2022gmflow}
Haofei Xu, Jing Zhang, Jianfei Cai, Hamid Rezatofighi, and Dacheng Tao.
\newblock Gmflow: Learning optical flow via global matching.
\newblock In \emph{Proceedings of the IEEE/CVF conference on computer vision
  and pattern recognition}, 2022.

\bibitem[Xu et~al.(2025)Xu, Li, Dong, Zhou, Newcombe, and Lv]{xu20254dgt}
Zhen Xu, Zhengqin Li, Zhao Dong, Xiaowei Zhou, Richard Newcombe, and Zhaoyang
  Lv.
\newblock 4dgt: Learning a 4d gaussian transformer using real-world monocular
  videos.
\newblock \emph{arXiv preprint arXiv:2506.08015}, 2025.

\bibitem[Yang et~al.(2021{\natexlab{a}})Yang, Sun, Jampani, Vlasic, Cole,
  Chang, Ramanan, Freeman, and Liu]{Yang2021LASRLA}
Gengshan Yang, Deqing Sun, V. Jampani, Daniel Vlasic, Forrester Cole, Huiwen
  Chang, Deva Ramanan, William~T. Freeman, and Ce Liu.
\newblock Lasr: Learning articulated shape reconstruction from a monocular
  video.
\newblock \emph{2021 IEEE/CVF Conference on Computer Vision and Pattern
  Recognition (CVPR)}, 2021{\natexlab{a}}.

\bibitem[Yang et~al.(2021{\natexlab{b}})Yang, Sun, Jampani, Vlasic, Cole, Liu,
  and Ramanan]{NEURIPS2021_a11f9e53}
Gengshan Yang, Deqing Sun, Varun Jampani, Daniel Vlasic, Forrester Cole, Ce
  Liu, and Deva Ramanan.
\newblock Viser: Video-specific surface embeddings for articulated 3d shape
  reconstruction.
\newblock In \emph{Advances in Neural Information Processing Systems}. Curran
  Associates, Inc., 2021{\natexlab{b}}.

\bibitem[Yang et~al.(2021{\natexlab{c}})Yang, Vo, Neverova, Ramanan, Vedaldi,
  and Joo]{Yang2021BANMoBA}
Gengshan Yang, Minh Vo, Natalia Neverova, Deva Ramanan, Andrea Vedaldi, and
  Hanbyul Joo.
\newblock Banmo: Building animatable 3d neural models from many casual videos.
\newblock \emph{2022 IEEE/CVF Conference on Computer Vision and Pattern
  Recognition (CVPR)}, 2021{\natexlab{c}}.

\bibitem[Yang et~al.(2023)Yang, Gao, Li, Gao, Wang, and Zheng]{yang2023track}
Jinyu Yang, Mingqi Gao, Zhe Li, Shang Gao, Fangjing Wang, and Feng Zheng.
\newblock Track anything: Segment anything meets videos.
\newblock \emph{arXiv preprint arXiv:2304.11968}, 2023.

\bibitem[Yang et~al.(2024{\natexlab{a}})Yang, Kang, Huang, Xu, Feng, and
  Zhao]{depthanything}
Lihe Yang, Bingyi Kang, Zilong Huang, Xiaogang Xu, Jiashi Feng, and Hengshuang
  Zhao.
\newblock Depth anything: Unleashing the power of large-scale unlabeled data.
\newblock In \emph{Proceedings of the IEEE/CVF conference on computer vision
  and pattern recognition}, 2024{\natexlab{a}}.

\bibitem[Yang et~al.(2024{\natexlab{b}})Yang, Gao, Zhou, Jiao, Zhang, and
  Jin]{yang2023deformable3dgs}
Ziyi Yang, Xinyu Gao, Wen Zhou, Shaohui Jiao, Yuqing Zhang, and Xiaogang Jin.
\newblock Deformable 3d gaussians for high-fidelity monocular dynamic scene
  reconstruction.
\newblock In \emph{Proceedings of the IEEE/CVF conference on computer vision
  and pattern recognition}, 2024{\natexlab{b}}.

\bibitem[Yang et~al.(2024{\natexlab{c}})Yang, Yang, Pan, Zhu, and
  Zhang]{yang2023real}
Zeyu Yang, Hongye Yang, Zijie Pan, Xiatian Zhu, and Li Zhang.
\newblock Real-time photorealistic dynamic scene representation and rendering
  with 4d gaussian splatting.
\newblock \emph{ICLR}, 2024{\natexlab{c}}.

\bibitem[Ye et~al.(2025)Ye, Li, Kerr, Turkulainen, Yi, Pan, Seiskari, Ye, Hu,
  Tancik, et~al.]{ye2024gsplatopensourcelibrarygaussian}
Vickie Ye, Ruilong Li, Justin Kerr, Matias Turkulainen, Brent Yi, Zhuoyang Pan,
  Otto Seiskari, Jianbo Ye, Jeffrey Hu, Matthew Tancik, et~al.
\newblock gsplat: An open-source library for gaussian splatting.
\newblock \emph{Journal of Machine Learning Research}, 26\penalty0 (34), 2025.

\bibitem[Yoon et~al.(2020)Yoon, Kim, Gallo, Park, and Kautz]{yoon2020novel}
Jae~Shin Yoon, Kihwan Kim, Orazio Gallo, Hyun~Soo Park, and Jan Kautz.
\newblock Novel view synthesis of dynamic scenes with globally coherent depths
  from a monocular camera.
\newblock In \emph{Proceedings of the IEEE/CVF Conference on Computer Vision
  and Pattern Recognition}, 2020.

\bibitem[Zhang et~al.(2024)Zhang, Herrmann, Hur, Jampani, Darrell, Cole, Sun,
  and Yang]{zhang2024monst3r}
Junyi Zhang, Charles Herrmann, Junhwa Hur, Varun Jampani, Trevor Darrell,
  Forrester Cole, Deqing Sun, and Ming-Hsuan Yang.
\newblock Monst3r: A simple approach for estimating geometry in the presence of
  motion.
\newblock \emph{arXiv preprint arxiv:2410.03825}, 2024.

\bibitem[Zhang et~al.(2021)Zhang, Cole, Tucker, Freeman, and
  Dekel]{zhang2021consistent}
Zhoutong Zhang, Forrester Cole, Richard Tucker, William~T Freeman, and Tali
  Dekel.
\newblock Consistent depth of moving objects in video.
\newblock \emph{ACM Transactions on Graphics (TOG)}, 40\penalty0 (4), 2021.

\bibitem[Zhang et~al.(2022)Zhang, Cole, Li, Rubinstein, Snavely, and
  Freeman]{zhang2022structure}
Zhoutong Zhang, Forrester Cole, Zhengqi Li, Michael Rubinstein, Noah Snavely,
  and William~T Freeman.
\newblock Structure and motion from casual videos.
\newblock In \emph{European Conference on Computer Vision}. Springer, 2022.

\bibitem[Zheng et~al.(2023)Zheng, Harley, Shen, Wetzstein, and
  Guibas]{zheng2023pointodyssey}
Yang Zheng, Adam~W Harley, Bokui Shen, Gordon Wetzstein, and Leonidas~J Guibas.
\newblock Pointodyssey: A large-scale synthetic dataset for long-term point
  tracking.
\newblock In \emph{Proceedings of the IEEE/CVF International Conference on
  Computer Vision}, 2023.

\bibitem[Zollh{\"o}fer et~al.(2014)Zollh{\"o}fer, Nie{\ss}ner, Izadi, Rhemann,
  Zach, Fisher, Wu, Fitzgibbon, Loop, Theobalt, and
  Stamminger]{Zollhfer2014RealtimeNR}
Michael Zollh{\"o}fer, Matthias Nie{\ss}ner, Shahram Izadi, Christoph Rhemann,
  Christopher Zach, Matthew Fisher, Chenglei Wu, Andrew~William Fitzgibbon,
  Charles~T. Loop, Christian Theobalt, and Marc Stamminger.
\newblock Real-time non-rigid reconstruction using an rgb-d camera.
\newblock \emph{ACM Transactions on Graphics (TOG)}, 33, 2014.

\end{thebibliography}
}
\clearpage
\appendix
\section*{Appendix}

\section{Additional Preprocessing Details}
\noindent\textbf{Obtaining Camera Poses.}
Our method takes video sequences with known camera poses as input. To obtain camera poses for in-the-wild videos with moving objects, we adopt one of these two approaches depending on the type of input camera motion: (1) if there is sufficient camera motion parallax, we use COLMAP~\cite{schoenberger2016sfm}'s SfM pipeline to obtain the camera poses and the sparse point clouds for the static regions, where we exclude keypoints in the foreground masks during the feature extraction stage.  
The foreground masks are generated using Track-Anything~\cite{yang2023track}, a flexible and interactive tool for video object tracking and segmentation. 
The static point clouds produced by COLMAP~\cite{schoenberger2016sfm} can then be used for aligning the affine-invariant monocular depth maps from Depth Anything~\cite{depthanything}. 
(2) If the video is captured by a roughly stationary camera (small interframe camera baseline), COLMAP tends to fail catastrophically. Initial results on DAVIS use camera poses estimated with Unidepth~\cite{piccinelli2024unidepth} and DroidSLAM~\cite{teed2021droid}. Specifically, we first employ Unidepth to predict metric depth and camera intrinsics. Using these predictions, we then estimate camera poses with DroidSLAM. (3) Camera parameters of in-the-wild videos in supplemental videos are estimated with MegaSaM~\cite{li2024_MegaSaM}.

\vspace{1em}
\noindent\textbf{Aligning Monocular Depth Maps.} The dispairty output from Depth Anything~\cite{depthanything} model is affine-invariant, so we need to align them with the cameras and reconstruction. To do so, we solve for a per-frame global scale and shift parameters that minimizes $\ell_2$ distance between the monocular disparity from Depth Anything and the disparity derived from SfM sparse point clouds or depth outputs from DroidSLAM or MegaSaM.

\vspace{1em}
\noindent\textbf{Computing Long-Range 2D Tracks.}
We utilize TAPIR~\cite{doersch2023tapir} to compute long-range 2D tracks for a video. While the ideal scenario would involve computing full-length tracks for every pixel in every frame (i.e., exhaustive pairs of correspondences), this approach is prohibitively computationally expensive.
We therefore only compute full-length tracks for pixels located on a grid for foreground moving objects in each frame.  In our experiment, we set grid interval to $4$~(i.e., sampling a query point every 4 pixels).
Due to our low-dimensional motion representation, we observe that our method can effectively operate with semi-dense 2D tracks without significant performance degradation 
During training, we filter out correspondences that TAPIR predicts with high uncertainty or those that are occluded.

\section{Additional Training Details}
\subsection{Initialization Details}
During initialization stage, we first solve a Procrustes alignment problem for each cluster $b$, where we estimate $\SETHREE$ transformation between point sets $\{\tf X_0\}_b$ 
and $\{\tf X_\tau\}_b$ for all $\tau = 0, \dots, T$ . We exclude point pair when either one of them is occluded, and weight the point pair using the uncertainty score predicted by TAPIR~\cite{doersch2023tapir} when solving Procrustes. 
This process produces the initialization for the set of basis functions $\{ \tf{T}_{0 \rightarrow t}^{(b)} \}_{b=1}^{B}$. 
To initialize the motion coefficient $\tf w^{(b)}$ for each track, we compute the distance between the 3D location of the track in the canonical frame and the 3D location of each cluster center, and initialize each of the corresponding motion coefficient value to be exponentially decay with the distance. 

We then optimize the $\gmean_0$, $\tf w^{(b)}$, and set of basis functions $\{ \tf{T}_{0 \rightarrow t}^{(b)} \}_{b=1}^{B}$ to fit the observed 3D tracks lifted by monocular depths and 2D tracks. Specifically, we enforce an  $\ell_1$ loss between each of our predicted 3D track and its corresponding observed 3D track. In addition, we enforce the motion bases to be temporally smooth by adding an $\ell_2$ regularization on the acceleration of both the quaternion and the translation vector. We optimize the parameters using Adam~\cite{kingma2014adam} optimizer for $2$k steps, where the initial learning rates for $\gmean_0$, $\tf w^{(b)}$, and $\{ \tf{T}_{0 \rightarrow t}^{(b)} \}_{b=1}^{B}$ are $1\times10^{-3}$, $1\times10^{-2}$ and $1\times10^{-2}$, respectively. All learning rates are exponentially decayed to $\frac{1}{10}$ of their initial values during the optimization process.

\subsection{Training Details}
\noindent\textbf{Gaussian Initialization.} The aforementioned initialization stage gives the initialization of the mean of each Gaussian in the canonical frame. We follow the original 3D-GS~\cite{kerbl3Dgaussians} paper to initialize the scale $\mathbf{s}$, rotation $\mathbf{R}_0$, and opacity $o$ of each Gaussian in the canonical frame. The color $\mathbf{c}$ of each Gaussian is initialized as the pixel color at the projected location in the canonical frame. 

\vspace{1em}
\noindent\textbf{Optimization.} We use Adam~\cite{kingma2014adam} Optimizer to optimize all scene parameters. The learning rates for Gaussian's canonical mean~ $\mathbf{\mu}_0$, opacity~$o$, scale~$\mathbf{s}$, rotation~$\mathbf{R}_0$~(parameterized as quaternion) and color $\mathbf{c}$ are set to $1.6\times10^{-4}$, $1\times10^{-2}$, $5\times10^{-3}$, $1\times10^{-3}$, and $1\times10^{-2}$, respectively. The learning rates for the $\SETHREE$ motion bases and the motion coefficients are set to $1.6\times10^{-4}$ and $1\times10^{-2}$ respectively. 
During each training iteration, we randomly select a batch of 8 query frames.
For each query frame, we render the color, mask, depth, and the 3D track locations for 4 randomly selected target frames.

\vspace{1em}
\noindent \textbf{Loss weights and details.} Our first set of loss function enforces our rendered results to match the per-frame pixelwise color, depth, and masks inputs.
The coefficients for depth loss  $\lambda_{\text{depth}}$ and mask loss $\lambda_{\text{mask}}$ are set to $0.5$ and $1.0$, respectively.
We additionally add regularization to the estimated surface geometry via a depth gradient loss and per-Gaussian scale penalty. In particular,
We add a pixelwise $\ell_1$ loss of spatial gradient between the depth renderings $\hat{\mathbf{D}}_t$ and corresponding reference depth maps $\mathbf{D}$, with a weight set to 1.0.
In addition, we enforce the foreground Gaussian to be isotropic by incorporating a regularization term that penalizes standard deviations of the scale $\mathbf{s}$ along all three axes.

Our second set of losses supervise the motion of the Gaussian.
The weights for the 2D tracking loss $\lambda_{\text{track-2d}}$ and the track depth loss $\lambda_{\text{track-depth}}$ are set to $2.0$ and $0.1$, respectively. 
The $L_{\text{track-2d}}$ is applied on normalized pixel coordinates~(i.e., divided by the maximum image edge length).
For the rigidity loss~$L_{\text{rigidity}}$, we use foreground part masks from SAM automatic segmentation independently on each frame.
For each training iteration, we sample 4 part masks for each query frame.
For each mask, we sample 32 center points and find their 16 nearest neighbors within the mask.
We compute the 3D track locations for all point samples in 4 target frames,
and regularize the distances between the center points and their neighbors in the target frames to be similar to their distances in the query frame.
We let $\lambda_{\textrm{rigidity}} = 0.1$.
We weight the neighbor distances with an exponential kernel $\exp ( -\beta \| x - \bar{x} \|)$ with $\beta = 2$, where $\bar x$ is the center point and $x$ is one of its neighbors.
We additionally add motion smoothness regularization that enforces an $\ell_2$ loss on acceleration of the motion translation bases and motion quaternion bases, with a weight set to 0.1 and an $\ell_2$ loss on the acceleration along $z$-axis (in the camera frame) of the $\mu_t$, both through finite difference approximation.

\vspace{1em}
\noindent \textbf{Training with 2D Gaussian Splatting (2DGS)} To enhance scene geometry estimation, we replace 3D Gaussian Splatting with 2D Gaussian Splatting~\cite{Huang2DGS2024}. Specifically, we employ an off-the-shelf monocular normal estimator~\cite{he2024lotusdiffusionbasedvisualfoundation} to generate monocular normal maps, $\tilde{N}$, which are then used to supervise both the rendered normals and the normals derived from the rendered depth. The supervision is achieved through the following losses:
\begin{equation}
    L_n = \sum_{i}\omega_{i}(1 - n_{i}^{T}\tilde{N})
\end{equation}

\begin{equation}
    L_N = 1 - N^{T}\tilde{N}
\end{equation}
where the summation is over the splats intersect the current ray, $\omega_{i}$ represents the blending weight for $i$-th splat, and $n_{i}$ denotes the normal of $i$-th splat. Here, $N$ represents the normal derived from rendered depth map. This supervision ensures that the orientation and depth of the 2D splats are aligned with the monocular normal prediction.

\section{Additional Evaluation Details}
In our evaluation, since the synthetic Kubric dataset~\cite{greff2022kubric} comes with groundtruth camera poses, we directly use the groundtruth camera poses for our experiments. For iPhone dataset~\cite{gao2022dynamic}, we observed that the provided camera poses from ARKit are not accurate, we thus perform an additional global bundle adjustment using COLMAP~\cite{schoenberger2016sfm} to refine the camera poses while fixing the camera intrinsics. To maintain metric scale after refinement, we compute a global $\mathbb{SIM(}3)$ transformation for each scene to align the refined camera poses with the original metric-scale camera poses. This allows us to evaluate 3D tracking performance in metric scale.

\section{Visualization of Kubric Experiment}
We find qualitatively that the optimized motion coefficients of the scene representation
are coherently grouped with each moving object in the scene.
We demonstrate this in Figure~\ref{fig:kubric},
where we show the first 3 PCA components of our optimized motion coefficients of evaluation scenes.

\providecommand\animage{}
\renewcommand{\animage}[2]{
    \frame{\includegraphics[width=.9\linewidth,clip,trim=#1]{figures/kubric_pca/#2}}
}
\begin{figure}[htbp]
\begin{minipage}[b]{0.45\textwidth}
\setlength{\tabcolsep}{0.8pt}
    \renewcommand{\arraystretch}{0.5}
    \centering
    \footnotesize
    \begin{tabularx}{\textwidth}{@{}*{2}{C}@{}}
        \animage{0 0 0 0}{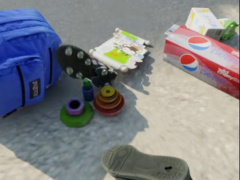} &
        \animage{0 0 0 0}{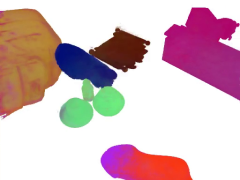}
        \\
        {Input} &
        {PCA Coef.}
    \end{tabularx}
    \vspace{-6pt}
    \caption{
        \textbf{First three PCA components of the optimized motion coefficients.}
    }
    \vspace{-1em}
    \label{fig:kubric}
\end{minipage}
\end{figure}

\section{{Visualization of Complex Dynamic Scene}}
\begin{figure}[h]
  \footnotesize
  \setlength{\abovecaptionskip}{1pt}
  \centering
  \includegraphics[width=\linewidth]{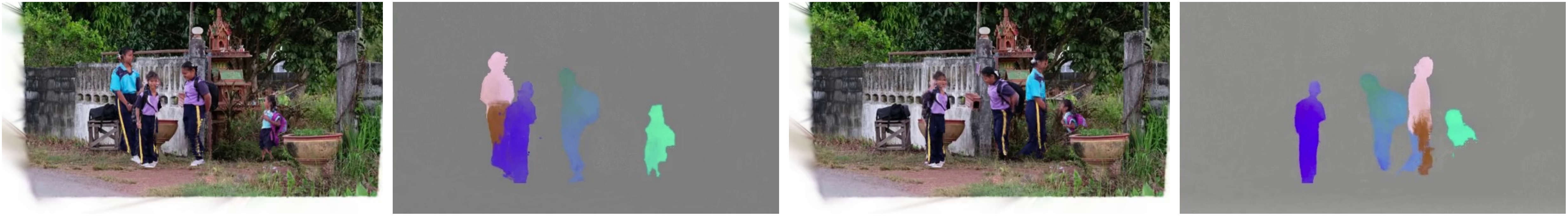}
  \caption{\footnotesize Novel view and motion coefficient PCA visualizations at time steps 0 and 54 of the \textit{school-girl} sequence from the DAVIS dataset.
  }
  \label{fig:motion_coef}
  \vspace{-10pt}
\end{figure}

{Our method is able to handle scenes with multiple moving objects. We visualize novel views and motion PCA of a real world complex scene in Fig.~\ref{fig:motion_coef}}.

\vspace{-0.5em}

\section{NVIDIA Dataset Evaluation} We conduct experiments on seven scenes from the NVIDIA Dynamic Scenes dataset \cite{yoon2020novel}, following the evaluation protocol of Dynamic Gaussian Marbles \cite{stearns2024dynamic}. Specifically, we use video footage from the static camera 4 for training and employ videos from cameras 3, 5, and 6 for evaluation. For consistency with Dynamic Gaussian Marbles \cite{stearns2024dynamic}'s experiments, we use images at half-resolution for all seven experiments, though our method is compatible with high-resolution images as well. Camera poses are estimated using COLMAP \cite{schoenberger2016sfm}, and depths are predicted with Depth Anything \cite{depthanything} and subsequently aligned with the COLMAP point cloud. The quantitative results reported in this paper differ from those in DGM~\cite{stearns2024dynamic}. Specifically, we compute covisibility masks between training and test views and apply them during evaluation, whereas DGM~\cite{stearns2024dynamic} inpaints unobserved regions in test views for evaluation. This difference in evaluation procedure accounts for the variation in reported performance. We will make the covisibility masks publicly available to facilitate future research.

\section{DynMF Implementation Detail}
DynMF ~\cite{kratimenos2024dynmf} shares similar design choices with our method. Since they don't have public code available, we implement our own version of it and provide implementation details here. Follow equation (4) in DynMF~\cite{kratimenos2024dynmf}, each motion basis is represented as a MLP:
\begin{equation}
    b_j(t) = MLP_j\left(\frac{t}{T}\right)
\end{equation}
where $b_j$ is the $j-th$ motion basis, $t$ is the queried time step, and $T$ is the number of time frames. We use a total of 10 motion bases. We apply positional encoding with 10 frequency bandwidths~\cite{mildenhall2020nerf} to the input $\frac{t}{T}$. Following DynMF~\cite{kratimenos2024dynmf}, we apply displacement motion to Gaussian's means and rotation quaternions as follows:
\begin{equation}
    \mu(t) = \mu_c + \sum_{j=1}^{10}c_jb_j^\mu(t)
\end{equation}
where $\mu_c$ is the Gaussian means in the canonical frame and $\{b_j^\mu(t)\}_{j=1}^{10}$ are the mean motion bases represented as MLPs.\\
\noindent and
\begin{equation}
    q(t) = q_c + \sum_{j=1}^{10}c_jb_j^R(t)
\end{equation}
where $q_c$ is the Gaussian rotation represented as quaternion in the canonical frame and $\{b_j^R(t)\}_{j=1}^{10}$ are the rotation motion bases represented as MLPs.The motion coefficients $\{c_j\}_{j=1}^{10}$ are shared between mean and rotation motions.

\end{document}